\pdfoutput=1

\documentclass[11pt]{article}
\usepackage[]{acl}

\usepackage{times}
\usepackage{latexsym}

\usepackage[T1]{fontenc}

\usepackage[utf8]{inputenc}

\usepackage{microtype}

\usepackage{inconsolata}

\usepackage{graphicx}
\usepackage{tabularx} 
\usepackage{inconsolata}
\usepackage{pifont}
\usepackage{algorithm}
\usepackage{algorithmic}
\usepackage[namelimits]{amsmath} 
\usepackage{amssymb}             
\usepackage{amsfonts}            
\usepackage{mathrsfs}            
\usepackage{graphicx} 
\usepackage{subfigure} 
\usepackage{xcolor}
\definecolor{darkgreen}{HTML}{3CB371}
\definecolor{darkred}{HTML}{7B7B7B}
\definecolor{ggreen}{HTML}{d4e7cf}
\definecolor{bblue}{HTML}{00bfff}
\definecolor{ppink}{HTML}{ff69b4}

\usepackage{makecell}
\usepackage{booktabs}
\usepackage{multirow}
\usepackage{enumitem} 
\usepackage{CJKutf8}

\usepackage{newfloat}
\usepackage{listings}
\usepackage{subcaption}
\usepackage{xspace}
\usepackage{graphicx}
\usepackage{caption}

\usepackage{amsmath}
\usepackage{mathtools}

\definecolor{green}{RGB}{36, 214, 36}
\definecolor{red}{RGB}{235, 30, 30}
\newcommand{\cmark}{\ding{51}} 
\newcommand{\xmark}{\ding{55}} 
\usepackage[utf8]{inputenc}
\usepackage[T1]{fontenc}

\usepackage{CJKutf8}
\usepackage{booktabs} 

\usepackage{CJK}
\usepackage{xcolor}
\usepackage{ulem}

\usepackage{booktabs}
\usepackage{tcolorbox}

\usepackage{soul}
\newcommand{\kaiti}[1]{\begin{CJK*}{UTF8}{gkai} #1 \end{CJK*}}
\soulregister{\kaiti}7
\definecolor{MyYellow}{rgb}{254, 246, 170}
\definecolor{MyBlue}{rgb}{170, 217, 251}

\usepackage{color}
\usepackage{colortbl}
\definecolor{g1}{RGB}{232,232,232}
\definecolor{g2}{RGB}{207,207,207}
\definecolor{g3}{RGB}{181,181,181}
\definecolor{g4}{RGB}{156,156,156}

\definecolor{b1}{RGB}{217,217,254}
\definecolor{b2}{RGB}{198,198,253}
\definecolor{b3}{RGB}{180,180,252}
\definecolor{b4}{RGB}{162,162,252}
\definecolor{b5}{RGB}{136,136,255}

\definecolor{r1}{RGB}{254,236,236}
\definecolor{r2}{RGB}{254,217,217}
\definecolor{r3}{RGB}{253,198,198}
\definecolor{r4}{RGB}{253,180,180}
\definecolor{r5}{RGB}{252,128,127}

\title{Edit Once, Update Everywhere: A Simple Framework for Cross-Lingual Knowledge Synchronization in LLMs}

\author{%
  Yuchen Wu$^{1}$,
  Liang Ding$^{2}$\thanks{Correspond to Liang Ding \texttt{liangding.liam@gmail.com}},
  Li Shen$^{3}$,
  Dacheng Tao$^{4}$\\
  $^{1}$Shanghai Jiao Tong University, China 200240\\
  $^{2}$The University of Sydney, Australia 2006\\
  $^{3}$Shenzhen Campus of Sun Yat-sen University, China 518107\\
  $^{4}$Nanyang Technological University, Singapore 639798 
  }

\begin{document}
\maketitle
\begin{abstract}
Knowledge editing allows for efficient adaptation of large language models (LLMs) to new information or corrections without requiring full retraining. However, prior methods typically focus on either single-language editing or basic multilingual editing, failing to achieve true cross-linguistic knowledge synchronization. To address this, we present a simple and practical state-of-the-art (SOTA) recipe \textit{Cross-Lingual Knowledge Democracy Edit} (X-KDE), designed to propagate knowledge from a dominant language to other languages effectively. Our X-KDE comprises two stages: (i) Cross-lingual Edition Instruction Tuning (XE-IT), which fine-tunes the model on a curated parallel dataset to modify in-scope knowledge while preserving unrelated information, and (ii) Target-language Preference Optimization (TL-PO), which applies advanced optimization techniques to ensure consistency across languages, fostering the transfer of updates. Additionally, we contribute a high-quality, cross-lingual dataset, specifically designed to enhance knowledge transfer across languages. Extensive experiments on the Bi-ZsRE and MzsRE benchmarks show that X-KDE significantly enhances cross-lingual performance, achieving an average improvement of +8.19\%, while maintaining high accuracy in monolingual settings. Our code is available at: \url{https://github.com/YukinoshitaKaren/X_KDE}.

\end{abstract}
\section{Introduction}

Large Language Models (LLMs) \cite{achiam2023gpt, dubey2024llama, yang2024qwen2, guo2025deepseek} have shown strong capabilities in natural language understanding, generation, and reasoning~\cite{wei2022emergent, zhong2023can, peng2023towards, zhao2023survey}. However, as world knowledge evolves, LLMs need methods to update outdated information efficiently. Knowledge editing \cite{yao2023editing} allows modifications to specific knowledge while preserving unrelated information, making it more cost-effective than retraining from scratch.

\begin{figure}[!t]
\centering
\hspace*{-14pt} %
\includegraphics[width=1.1\linewidth]{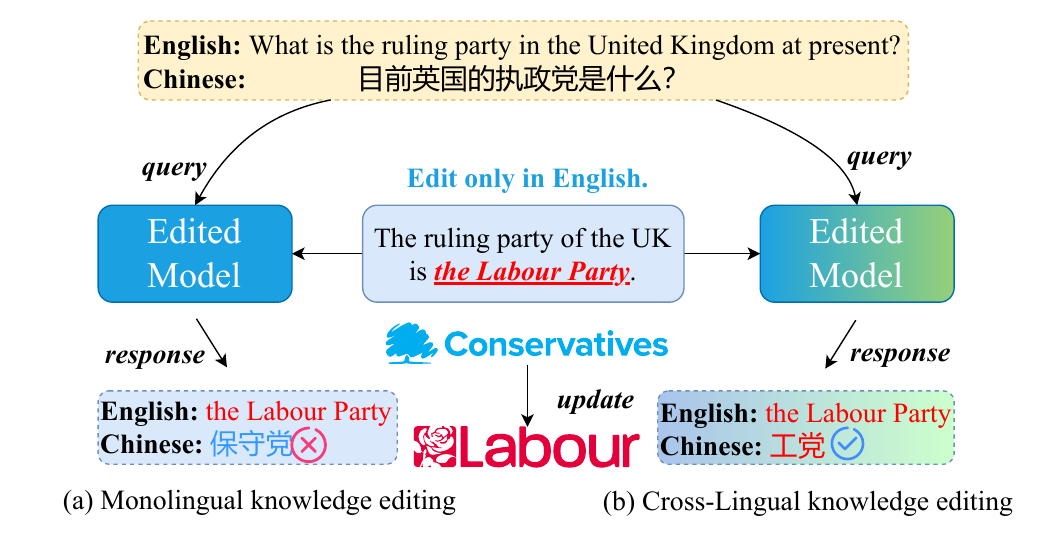}
\caption{Examples of (a) \textbf{monolingual} and (b) \textbf{cross-lingual} knowledge editing. In the former, the editing and verification languages are the same, while in the latter, knowledge is transferred from the source language (e.g., English) to the target language (e.g., Chinese).
}
\vspace{-10pt}
\label{fig:intro}
\end{figure}
Despite significant progress, most existing approaches focus on monolingual editing~\cite{de2021editing,dai2021knowledge,mitchell2021fast}. As LLMs are increasingly required to handle multilingual queries \cite{zhang2024comprehensive, wang2023towards}, monolingual solutions often fail. For example, when editing the response "the Conservative Party" to "the Labor Party" in English (Figure~\ref{fig:intro}(a)), this update does not propagate to the Chinese version. Thus, expanding knowledge editing to cross-lingual settings is crucial to ensure that changes made in the source language are correctly applied to target languages.

Currently, several studies on multilingual knowledge editing have emerged \cite{xu2022language, wang2023retrieval, wei2024mlake, xie2024memla, liang2024multilingual}. Some of these methods extend the edited language from single to multiple, while others prescribe source-language answers as the ground truth for multilingual queries. Both strategies fall short of achieving true cross-lingual knowledge democratization. For example, although IKE was regarded as the state-of-the-art method in previous studies~\cite{wang2023cross,xie2024memla}, its performance on the Bi-ZsRE benchmark demonstrates significant limitations, achieving merely a 73.33 average score when editing in English. Unlike previous methods that attempt to forcefully correct LLM behaviors in both the source and target languages, we propose guiding LLMs to internalize knowledge from source language editing and apply it to target language queries. Our \textit{Cross-Lingual Knowledge Democracy Edit} (X-KDE) with Dual-Stage Refinement, where we use parallel language datasets to transfer knowledge from the source to the target language.

The X-KDE framework involves two phases: (i) Cross-lingual edition Instruction Tuning (XE-IT), where the source language editing descriptor is paired with target language queries to create a parallel dataset, guiding the model to answer in the target language while preserving unchanged knowledge. (ii) Target-language Preference Optimization (TL-PO), where we adopt the ORPO strategy~\cite{hong2403orpo}, further constrains cross-lingual knowledge, promoting the diffusion of updates from source to target languages, and achieving true knowledge democratization. Taking the Bi-ZsRE benchmark as an example, X-KDE outperforms others, achieving average scores of 91.04 and 88.49 when editing in English and Chinese, respectively. Our \textbf{contributions} are three-fold:
\begin{itemize}
    \item To tackle the scarcity of high-quality resources in cross-lingual knowledge editing, we introduce new datasets that fill gaps in existing resources, enhancing the reliability of knowledge transfer across languages.
    \item We propose X-KDE, a simple yet highly effective method for cross-lingual knowledge editing. This approach, based on a two-stage process, ensures robust knowledge generalization across languages.
    \item Through extensive experiments, we establish X-KDE as a new state-of-the-art (SOTA) solution for cross-lingual knowledge editing, demonstrating significant improvements in performance while preserving original knowledge and enhancing the portability of updates.
\end{itemize}

\section{Preliminary}
\subsection{Knowledge Editing}
Knowledge editing selectively modifies in-scope knowledge while preserving out-of-scope behavior. Given a base LLM $p_{\theta}$ and an \textit{edit descriptor} $\langle x_e, y_e \rangle$, where $x_e$ is the modification description and $y_e$ is the corresponding answer, the edited model should adhere to four key properties:

\paragraph{Reliability} evaluates accuracy on edit descriptors:
\begin{equation}
\mathbb{E}_{(x_e, y_e) \sim \mathcal{X}_e}\mathbf{1}[\mathop{\arg\max}_{y} p_{\theta}^*(y|x_e) = y_e]
\end{equation}

\paragraph{Generality} assesses the precision of semantically rephrased examples:
\begin{equation}
\mathbb{E}_{(x_e^{par}, y_e) \sim \mathcal{X}_e^{par}}\mathbf{1}[\mathop{\arg\max}_{y} p_{\theta}^*(y|x_e^{par}) = y_e]
\end{equation}

\paragraph{Locality} ensures that out-of-scope inputs remain unchanged:
\begin{equation}
\mathbb{E}_{(x_e, y_e) \sim \mathcal{O}_e}\mathbf{1}[ p_{\theta}^*(y|x_e) = p_{\theta}(y|x_e)]
\end{equation}

\paragraph{Portability} measures the ability to transfer updated knowledge to related queries:
\begin{equation}
\mathbb{E}_{(x_e, y_e) \sim \mathcal{I}_e}\mathbf{1}[\mathop{\arg\max}_{y} p_{\theta}^*(y|x_e) = y_e]
\end{equation}

\subsection{Cross-Lingual Knowledge Editing}
Cross-lingual knowledge editing extends monolingual knowledge editing by requiring a multilingual LLM $p_{m\theta}$ to propagate knowledge from a source language to a target language. Given an edit descriptor in the source language $\langle x_e^s, y_e^s \rangle$, the goal is to maximize:

\begin{equation}
\mathbb{E}_{(x_e^s, y_e^s) \sim \mathcal{X}_e^s
\atop
x_e^t = I^t({x}^s_e), y_e^t = I^t({y}^s_e)
}\mathbf{1}[\mathop{\arg\max}_{y} p_{m\theta}^*(y|x_e^t) = y_e^t]
\end{equation}

\begin{equation}
\mathbb{E}_{(x_e^s, y_e^s) \sim \mathcal{O}_e^s
\atop
x_e^t = I^t({x}^s_e), y_e^t = I^t({y}^s_e)
}\mathbf{1}[p_{m\theta}^*(y|x_t) = p_{m\theta}(y|x_t)]
\end{equation}

Here, $x_e^t, y_e^t$ are the edit descriptors in the target language $t$, and $I^t(\cdot)$ translates the source language input into the target language. Cross-lingual knowledge editing demands cross-lingual comprehension, ensuring that updates in the source language lead to consistent responses in the target language.

\section{Methodology}

\begin{figure*}[!t]
\centering
\includegraphics[width=\linewidth]{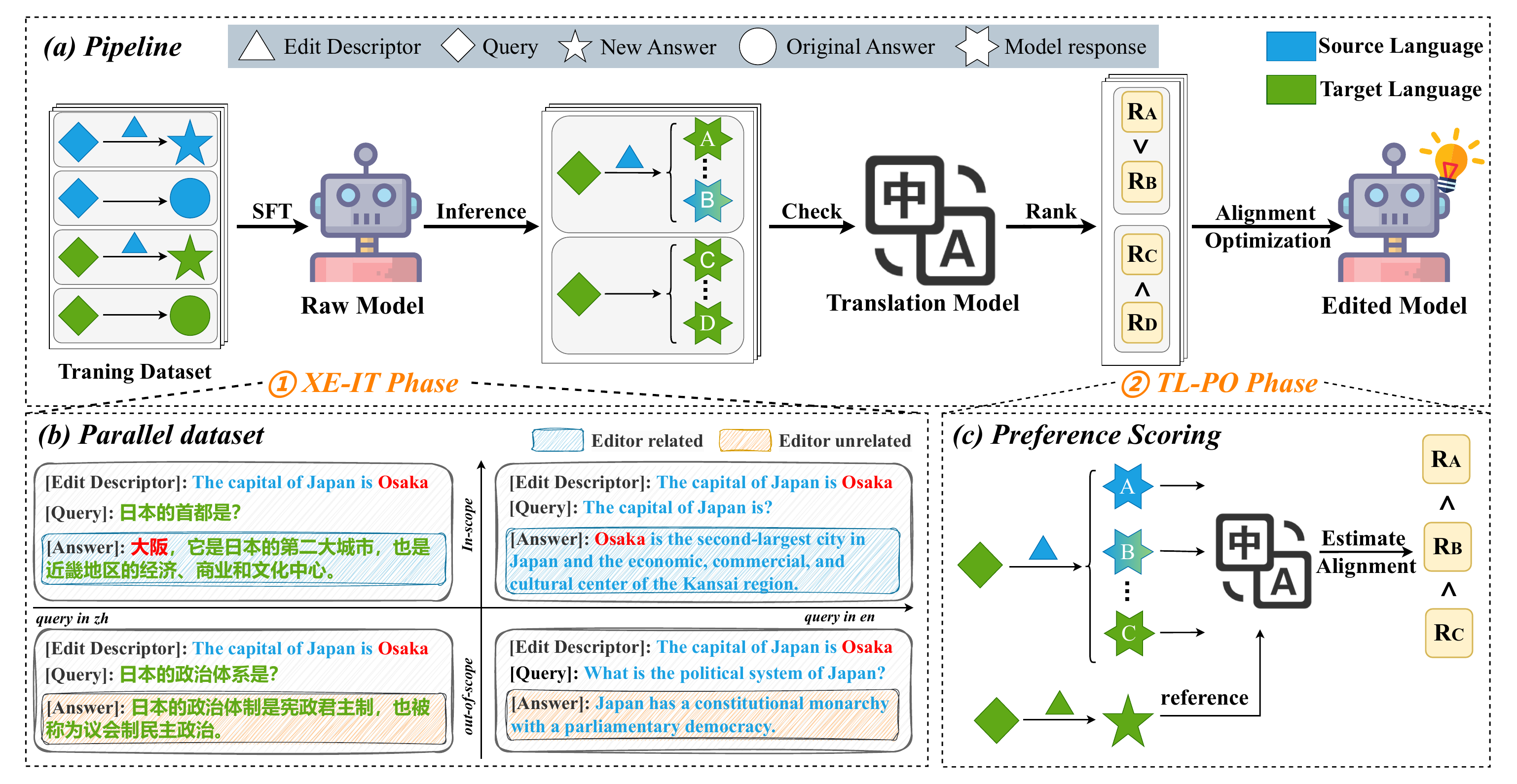}
\caption{\textbf{Illustration of \textit{Cross-Lingual Knowledge Democracy Edit} (X-KDE) framework}. In the XE-IT phase, we fine-tune the LLM on a carefully curated parallel dataset, enabling it to incorporate newly edited information from the source language when queried in the target language. In the TL-PO phase, multiple responses are generated, ranked based on similarity to the target language answer, and alignment optimization is applied to refine the output.}
\label{fig:method}
\end{figure*}

To achieve the democratization of knowledge, we propose the \textit{Cross-Lingual Knowledge Democracy Edit} (X-KDE) framework, as shown in Figure~\ref{fig:method}. This framework enables LLMs to adapt to evolving knowledge demands and facilitates the transfer of knowledge to target languages by editing only the source language. X-KDE consists of two stages: the Cross-lingual Edition Instruction Tuning (XE-IT) phase and the Target-language Preference Optimization (TL-PO) phase.

\subsection{XE-IT Stage: Learning to Edit}
\label{sec:XE-IT}

The goal is to enable the model to leverage knowledge edits in the source language while preserving the unchanged information. To meet the requirements for cross-lingual editing, we carefully constructed a high-quality dataset and employed XE-IT to fine-tune the model.

\subsubsection{Dataset Construction}

\paragraph{Data Sources.} Our goal is to enable the model to use edit descriptors effectively while maintaining unrelated information. We use several widely used knowledge editing datasets, including ZsRE~\cite{levy2017zero}, HalluEditBench~\cite{huang2024can}, RIPPLEEDITS~\cite{cohen2024evaluating}, WikiBio~\cite{hartvigsen2024aging}, MQUAKE~\cite{zhong2023mquake}, and COUNTERFACT~\cite{meng2022locating}, to build our training data. These datasets provide edit descriptors paired with QA pairs. 
Additionally, following LTE~\cite{jiang2024learning}, we incorporate Evol-Instruct~\cite{xu2023wizardlm} to better preserve the language capabilities of the LLM.
To mitigate data leakage, we randomly sample and translate subsets for training.

\paragraph{Sample Generation.} Existing datasets often feature straightforward QA pairs, which limit the model's comprehension ability. To address this, we use Deepseek~\cite{liu2024deepseek} to generate complex in-scope and out-of-scope query-answer pairs. This method enhances training data quality and model comprehension.

\paragraph{Quality Control.} To ensure relevance, we use Deepseek to assess the quality of in-scope query-answer pairs. Samples are scored based on Syntactic Structure, Lexical Richness, and Edit Consistency, with low-quality samples filtered out and replaced by higher-scoring ones.

\paragraph{Translation Process.} We use large language models, i.e., Deepseek\footnote{\url{https://api-docs.deepseek.com/}} to translate the generated data from English to Chinese.

\paragraph{Parallel Data Construction.} Our dataset follows a parallel structure (Figure~\ref{fig:method}(b)), with the in-scope section guiding LLMs when to use updated knowledge and the out-of-scope section minimizing the impact on unrelated knowledge. The dataset includes both monolingual and cross-lingual sections, where the source language contains edit descriptors and the target language contains queries and answers. Further details about our dataset are provided in Appendix~\ref{sec:appendix_data_construction}.

\subsubsection{Fine-Tuning}

Thanks to the flexible parallel structure, we can adaptively select the source and target languages to satisfy specific needs. We create a large-scale cross-lingual dataset and compute loss based only on the answer tokens. The model generates answers in the target language given source-language edit descriptors and target-language queries.

\subsection{TL-PO Stage: Preference Optimization}

\subsubsection{Preference Scoring}

After the XE-IT phase, the model has initially acquired the ability for cross-lingual knowledge editing. However, when faced with queries in the target language, the model may still make mistakes, such as generating responses in the source language, producing surface-level transliterations, or failing to follow target language patterns. To address this, we use a multilingual translation model to compute "alignment" scores, favoring responses aligned with the target language. More details about the "alignment" score computation can be found in Appendix~\ref{sec:aligenment}.

\subsubsection{Alignment Optimization}
When an edit is performed in the source language and the query is made in the target language, we aim for the model to generate answers in the target language with higher likelihood than in the source language: 
$p_{\theta}^*(y_e^t|x_e^s) > p_{\theta}^*(y_e^s|x_e^s)$. To achieve this, we employ ORPO, a state-of-the-art preference optimization method. We collect flawed outputs ($Y_l$) and preferred outputs ($Y_w$), then optimize the objective function:

\begin{equation}
    \mathcal{L}_{ORPO} = \mathbb{E}_{(x, y_w, y_l)}\left[ \mathcal{L}_{XE-IT} + \lambda \cdot \mathcal{L}_{OR} \right]
    \label{eq:main}
\end{equation}

where $\mathcal{L}_{XE-IT}$ is the XE-IT loss and $\mathcal{L}_{OR}$ maximizes the likelihood ratio between the preferred response and the less preferred one.

\begin{equation}
    \mathcal{L}_{XE-IT} = - \frac{1}{m} \sum_{t=1}^{m} \log P_\theta(y_t | x, y_{<t}) \label{eq:xe-it}
\end{equation}

\begin{equation}
    \mathcal{L}_{OR} = -\log \sigma \left( \log \frac{\textbf{odds}_\theta(y_w|x)}{\textbf{odds}_\theta(y_l|x)} \right) \label{eq:ratio} 
\end{equation}

\begin{equation}
    \textbf{odds}_\theta(y|x) = \frac{P_\theta(y|x)}{1 - P_\theta(y|x)}\label{eq:odds}
\end{equation}

\section{Experiments}
\subsection{Experimental Setup}

\paragraph{Baselines.} 

\begin{table*}[t]
    \centering
    \resizebox*{0.9\linewidth}{!}{
    \begin{tabular}{cccccccccc}
    \toprule
    \multicolumn{1}{c}{\multirow{2}{*}{\bf Method}} & \multicolumn{4}{c}{\textbf{Test in English}} & \multicolumn{4}{c}{\textbf{Test in Chinese}} & \\
    \cmidrule(lr){2-5} \cmidrule(lr){6-9} 
  \multicolumn{1}{c}{} & \textbf{Reliability} & \textbf{Generality} & \textbf{Locality} & \textbf{Portability} & \textbf{Reliability} & \textbf{Generality} & \textbf{Locality} & \textbf{Portability} & \textbf{\textit{\underline{Avg.}}} \\
    \midrule
\multicolumn{10}{c}{\textbf{Edit in English}} \\
\toprule
FT-L & 53.51 & 50.18 & 94.01 & 53.31 & 51.81 & 51.71 & 85.56 & 55.14 & \underline{61.90}\\
FT-M & 99.97 & 95.38 & 97.92 & 57.69 & 56.89 & 56.52 & 94.61 & 52.16 & \underline{76.39}\\
ROME & 96.09 & 84.69 & 98.04 & 58.87 & 49.94 & 50.31 & 97.70 & 51.81 & \underline{73.43}\\ 
MEMIT & 95.21 & 89.14 & \textcolor{darkgreen}{98.56} & 57.77 & 52.05 & 52.01 & \textcolor{darkgreen}{98.76} & 52.19 & \underline{74.46}\\  
IKE & 99.59 & 99.61 & 56.95 & 71.27 & 67.83 & 67.88 & 64.54 & 58.97 & \underline{73.33}\\ 
LTE & 99.91 & 99.81 & 88.97 & \textcolor{darkgreen}{77.40} & 76.86 & 76.82 & 86.99 & 67.49 & \underline{84.28}\\ 
\cmidrule{1-10}
X-KDE(Ours) & \textcolor{darkgreen}{99.93} & \textcolor{darkgreen}{99.87} & 90.15 & 76.41 & \textcolor{darkgreen}{94.81} & \textcolor{darkgreen}{94.65} & 95.05 & \textcolor{darkgreen}{77.43} & \underline{\textcolor{darkgreen}{91.04}} \\  
    \midrule
    
    \multicolumn{10}{c}{\textbf{Edit in Chinese}} \\
    \toprule
FT-L & 40.80 & 40.66 & 94.80 & 55.24 & 54.72 & 53.68 & 66.51 & 48.75 & \underline{56.89}\\
FT-M & 51.86 & 51.24 & 98.18 & 55.30 & \textcolor{darkgreen}{100.0} & 99.71 & 79.28 & 61.98 & \underline{74.69}\\
ROME & 44.14 & 43.80 & 97.92 & 52.66 & 72.24 & 70.12 & \textcolor{darkgreen}{96.48} & 48.15 & \underline{65.69}\\
MEMIT & 45.37 & 44.95 & \textcolor{darkgreen}{99.07} & 54.65 & 75.19 & 73.45 & 96.02 & 51.44 & \underline{67.52}\\ 
IKE & 65.87 & 65.74 & 69.41 & 63.06 & 99.86 & \textcolor{darkgreen}{99.73} & 64.86 & 72.39 &  \underline{80.79} \\ 
LTE & 64.63 & 62.56 & 85.23 & 62.6 & 99.79 & 99.31 & 87.17 & 69.69 &  \underline{78.87} \\
\cmidrule{1-10}
X-KDE(Ours) & \textcolor{darkgreen}{93.49} & \textcolor{darkgreen}{92.22} & 90.56 & \textcolor{darkgreen}{65.55} & \textcolor{darkgreen}{100.00} & 99.11 & 92.85 & \textcolor{darkgreen}{74.14} & \underline{\textcolor{darkgreen}{88.49}} \\
    \bottomrule
    \end{tabular}
}
    \caption{
    \textbf{Cross-lingual editing performance of different methods} on Llama2-chat-7B backbones. Results in \textcolor{darkgreen}{green} indicates the best results. ``\underline{\textbf{\textit{Avg.}}}'' represents the overall mean of all metrics evaluated across the two languages.
    }
    \label{table:F1-res}
\end{table*}

\begin{table*}[!h]
    \centering
    \scalebox{0.7}{
        \begin{tabular}{ccccccccccccccc}
        \toprule
         \textbf{Metrics} & \textbf{Methods}  & \textbf{en-en} & \textbf{en-cz} & \textbf{en-de} & \textbf{en-du} & \textbf{en-es} & \textbf{en-fr} & \textbf{en-pt} & \textbf{en-ru} & \textbf{en-th} & \textbf{en-tr} & \textbf{en-vi} & \textbf{en-zh} & \underline{\textbf{en-avg}}\\
         \midrule
           \multirow{7}{*}{{\textbf{Reliability}}} 
           & FT-L &  52.92 &  41.81 &  39.79 &  39.02 &  39.49 &  39.72 &  39.26 &  39.79 &  36.44 &  36.86 &  46.21 &  51.81 &  \underline{41.93} \\
           & FT-M &  99.96 &  66.93 &  70.16 &  67.17 &  63.69 &  64.98 &  64.22 &  48.96 &  36.46 &  57.54 &  66.80 &  56.89 &  \underline{63.65} \\
           & ROME  &  96.36 &  56.54 &  60.82 &  58.89 &  57.41 &  56.43 &  54.91 &  41.69 &  35.44 &  45.76 &  56.94 &  49.94 &  \underline{55.93} \\
           & MEMIT &  95.44 &  62.37 &  64.82 &  64.12 &  59.46 &  61.90 &  58.69 &  44.54 &  36.40 &  49.15 &  61.34 &  52.05 &  \underline{59.19} \\
           & IKE & 99.65 & 83.22 & 80.61 & 79.36 & 76.69 & 78.48 & 75.37 & 67.62 & 54.38 & 76.90 & 81.22 & 67.83 & \underline{76.78} \\
           & LTE  & \textcolor{darkgreen}{100.00} & 84.29 & 81.71 & 80.60 & 77.67 & 79.11 & 77.39 & 72.02 & 62.04 & 78.87 & 81.92 & 76.93 & \underline{79.38} \\
           \cmidrule{2-15}
           & X-KDE & 99.93 & \textcolor{darkgreen}{92.78} & \textcolor{darkgreen}{87.43} & \textcolor{darkgreen}{88.89} & \textcolor{darkgreen}{85.71} & \textcolor{darkgreen}{87.49} & \textcolor{darkgreen}{89.87} & \textcolor{darkgreen}{89.32} & \textcolor{darkgreen}{89.66} & \textcolor{darkgreen}{91.23} & \textcolor{darkgreen}{87.55} & \textcolor{darkgreen}{93.07} & \underline{\textcolor{darkgreen}{90.24}} \\
           \midrule
            \multirow{7}{*}{\textbf{Generality}} 
            & FT-L &  49.60 &  40.75 &  38.87 &  38.36 &  39.68 &  39.12 &  39.56 &  38.97 &  36.89 &  37.18 &  45.89 &  51.71 &  \underline{41.38} \\
            & FT-M &  95.53 &  65.45 &  68.15 &  65.09 &  62.39 &  62.28 &  61.63 &  47.69 &  36.88 &  56.87 &  65.97 &  56.52 &  \underline{62.04} \\
            & ROME &  85.13 &  54.99 &  58.91 &  56.99 &  56.58 &  54.47 &  53.94 &  40.68 &  35.36 &  45.06 &  56.38 &  50.31 &  \underline{54.07} \\
            & MEMIT &  89.59 &  60.71 &  63.80 &  61.98 &  58.10 &  59.40 &  57.63 &  43.31 &  36.77 &  48.68 &  60.51 &  52.01 &  \underline{57.71} \\
            & IKE & 99.54 & 82.67 & 80.78 & 79.18 & 76.37 & 78.22 & 75.49 & 67.51 & 54.26 & 76.97 & 80.99 & 67.88 & \underline{76.65} \\
            & LTE & \textcolor{darkgreen}{99.87} & 84.26 & 81.63 & 81.07 & 77.51 & 78.99 & 77.38 & 71.46 & 61.90 & 78.26 & 81.37 & 76.24 & \underline{79.16} \\
           \cmidrule{2-15}
           & X-KDE & 99.68 & \textcolor{darkgreen}{92.87} & \textcolor{darkgreen}{87.25} & \textcolor{darkgreen}{88.87} & \textcolor{darkgreen}{85.16} & \textcolor{darkgreen}{87.57} & \textcolor{darkgreen}{89.93} & \textcolor{darkgreen}{89.10} & \textcolor{darkgreen}{89.21} & \textcolor{darkgreen}{91.25} & \textcolor{darkgreen}{87.62} & \textcolor{darkgreen}{93.11} & \underline{\textcolor{darkgreen}{90.14}} \\
          
           \bottomrule
        \end{tabular}
    }
    \caption{
    \textbf{Results on MzsRE dataset for editing performed in English} using Llama2-7b-chat. Here, ``en-zh'' means that English serves as the source language and Chinese as the target language, with similar interpretations for the other pairs. \underline{``en-avg''} denotes the average performance across cross-lingual scenarios.}
    \label{tab:en-edit}
\end{table*}

We chose the following methods as baselines:
(1) \textbf{FT-L}~\cite{meng2022locating} fine-tunes a specific layer of the feed-forward network to maximize the likelihood of target tokens;
(2) \textbf{FT-M}~\cite{zhang2024comprehensive} fine-tunes the same feed-forward network layer as FT-L. Additionally, it masks the original text and applies cross-entropy loss on the target answer;
(3) \textbf{ROME}~\cite{meng2022locating} employs causal mediation analysis to identify the target area for editing, and then updates the parameters of the feed-forward network layers;
(4) \textbf{MEMIT}~\cite{meng2022mass}, built upon the ROME framework, enables the simultaneous update of thousands of knowledge;
(5) \textbf{IKE}~\cite{zheng2023can} utilizes the in-context-learning ability of the model and provides a few-shot demonstration to guide the model's responses based on the updated facts.
(6) \textbf{LTE}~\cite{jiang2024learning} enhances the model's instruction-following ability through supervised fine-tuning (SFT), and employs a retrieval-based mechanism to provide updated knowledge for demonstrations.

\paragraph{Dataset.}
We evaluate X-KDE and Baselines using two widely used benchmarks: Bi-ZsRE dataset~\cite{wang2023cross} and MzSRE dataset~\cite{wang2023retrieval}. The former contains 1,037 test samples in English and an equal number in Chinese, while the latter covers twelve languages: English, Czech, German, Dutch, Spanish, French, Portuguese, Russian, Thai, Turkish, Vietnamese and Chinese, with 743 samples in each.
\paragraph{Backbones.} 
We select two public models as backbones, including LLaMA2-Chat-7B~\cite{touvron2023llama} and Qwen2.5-instruct-7B~\cite{yang2024qwen2}. These models are widely used in chatbot applications, the former excels in English, while the latter demonstrates strong multilingual abilities. For brevity, the results on Qwen2.5-instruct-7B are provided in~\ref{sec:appendix_ex_qwen2}.

\begin{figure*}[!t]
\vspace{-10pt}
\centering
\includegraphics[width=1.0\linewidth]{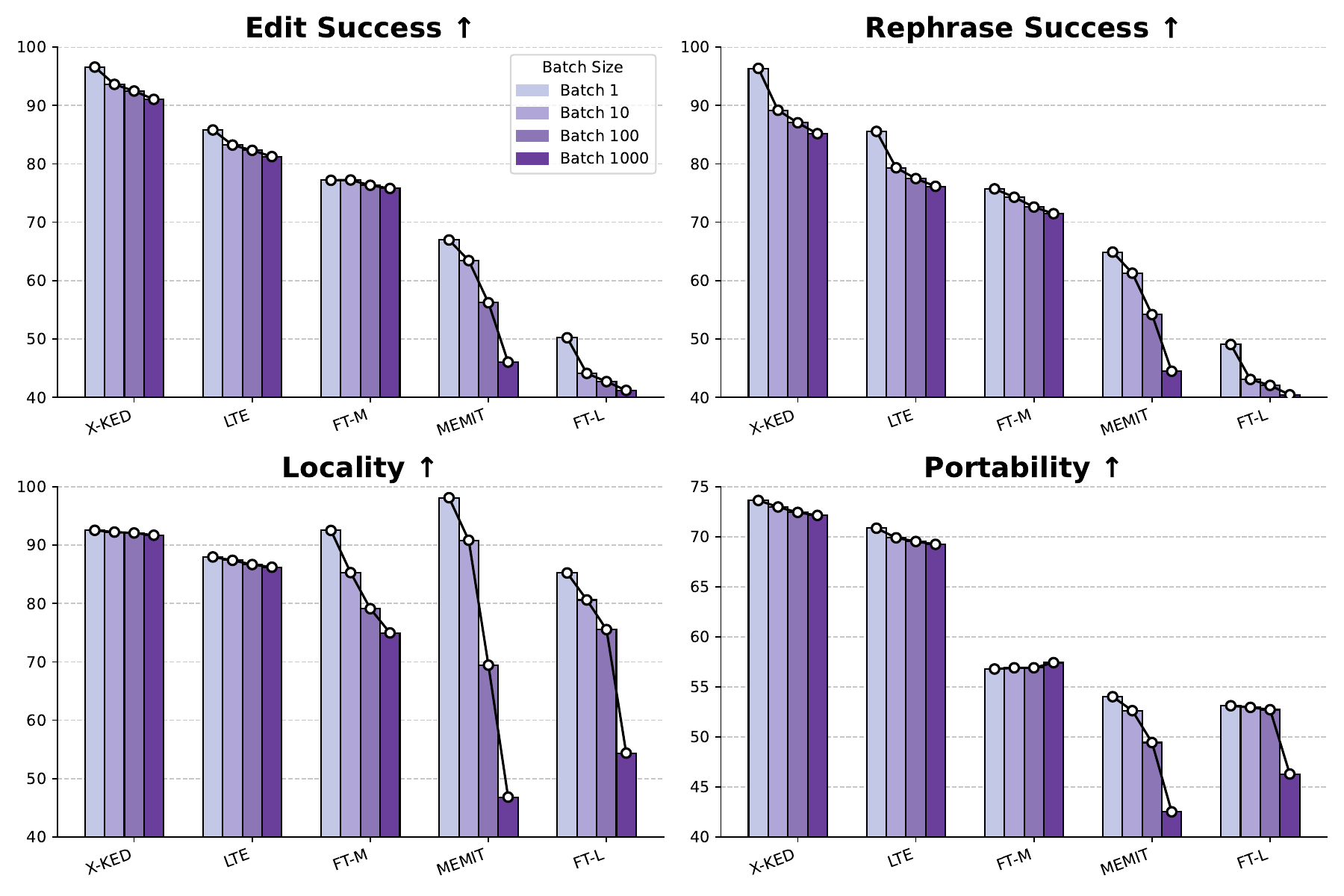}
\caption{
\textbf{Mean batch-editing performance across four benchmarks} at batch sizes 1, 10, 100, and 1000.
}
\label{fig:batch_edit}
\vspace{-10pt}
\end{figure*}

\begin{figure}[!t]
\centering
\includegraphics[width=1.0\linewidth]{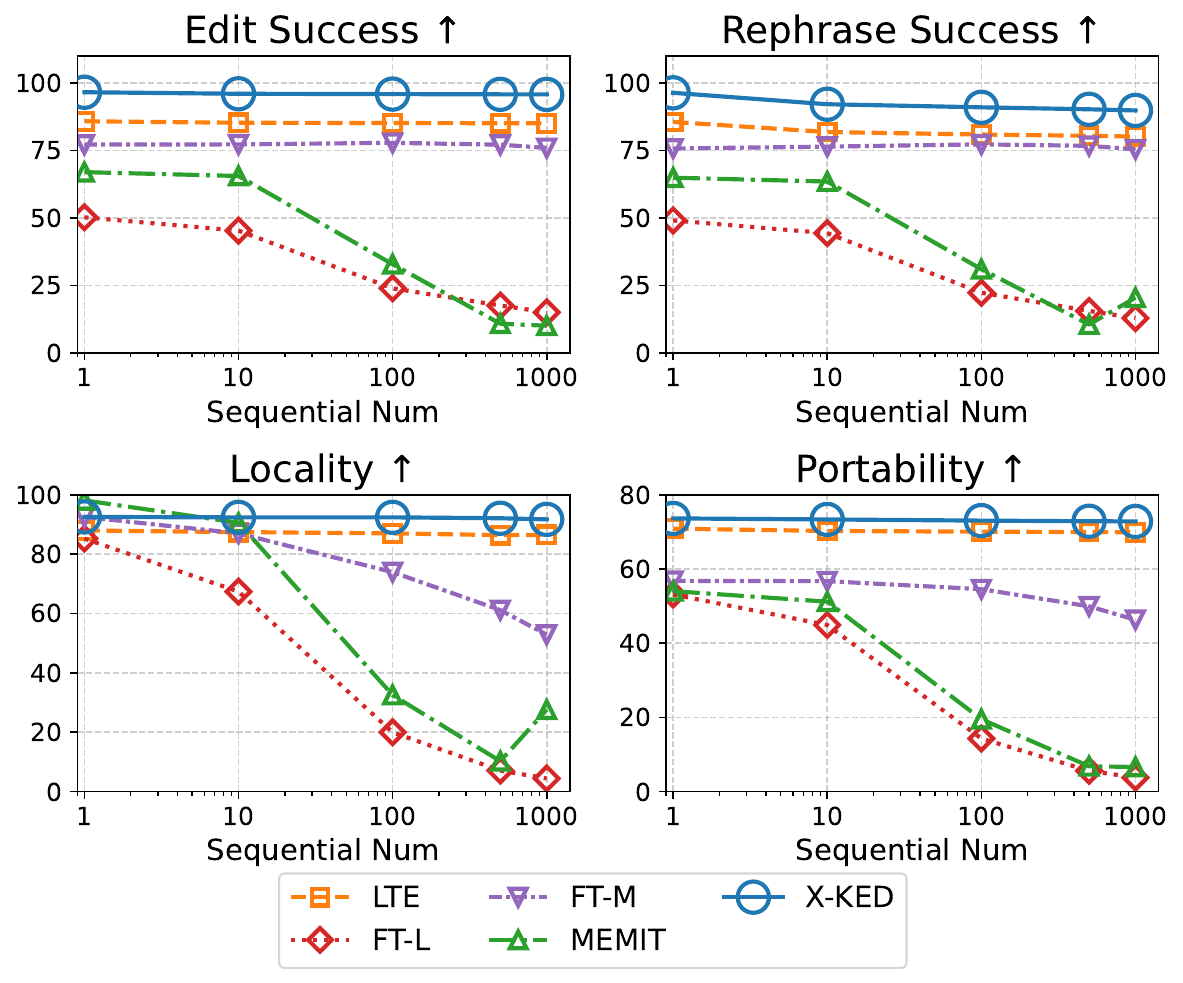}
\caption{
\textbf{Mean sequential-editing results across four knowledge editing benchmarks}, shown for data stream sizes of 1, 10, 100, 500, and 1000 (log-scale).
}
\label{fig:seq_edit}
\vspace{-10pt}
\end{figure}

\subsection{Results of Single Fact Editing} 
Table~\ref{table:F1-res} and Table~\ref{tab:en-edit} demonstrate the main results of single fact editing, which focus on single editing cases. From these results, we can find several significant observations:
\paragraph{X-KDE outperforms other methods in the cross-lingual setting by a significant margin.} 
As shown in Table~\ref{table:F1-res}, when edited in English, it is evident that our method brings average performance improvements of 6.76\%, compared to LTE.
In particular, in the cross-lingual setting, our method achieves further performance gains in portability, which demonstrates that our method not only captures surface-level changes in wording but enables the LLM to effectively internalize the knowledge edited in the source language and apply it to the target language. In summary, LTE sets a new state-of-the-art in cross-lingual knowledge editing task.

\paragraph{X-KDE brings consistent and effective improvements in more complex multilingual environments.} 
Our method is effective not only in a bilingual Chinese-English setting but can also be generalized to additional languages. We conducted more extensive experiments on the Mzsre dataset, and the results are presented in Table~\ref{tab:en-edit}. More detailed results in Appendix~\ref{sec:appendix_ex_llama2}.
It is evident that, compared to LTE, our method exhibits 10.86\% in reliability and 10.98\% average gains in generality when edit in English. 
These results further demonstrate that, our approach significantly enhances the model's cross-lingual abilities, enabling it to effectively apply knowledge from a pivot language to others and marking a significant step toward the democratization of knowledge.

\subsection{Results of Mass Fact Editing} 

In the previous section, we introduced the results of single fact editing. However, real-world scenarios are often more complex, requiring simultaneous or sequential edits to multiple pieces of knowledge. Therefore, in this section, we conduct comprehensive experiments using X-KDE alongside several methods that support mass editing (FT-L, FT-M, MEMIT, and LTE) on LLaMA2-Chat-7B in both batch-edit and sequential-edit settings, and then present the corresponding results.

\begin{table*}[t]
    \centering
    \resizebox{\linewidth}{!}{
    \begin{tabular}{ccccccccccc}
    \toprule
    
        & \multicolumn{5}{c}{\textbf{Test Language: en}} & \multicolumn{4}{c}{\textbf{Test Language: zh}} \\
        \cmidrule(lr){2-6} \cmidrule(lr){7-10} 
        &  \textbf{MMLU} & \textbf{CommonSenseQA} & \textbf{PIQA} & \textbf{Xsum} & \textbf{NQ} & \textbf{CMMLU} & \textbf{CommonSenseQA\_zh} & \textbf{CEval} & \textbf{NQ\_zh} & \textbf{\textit{avg}} \\
    \midrule
    \textit{LLaMA2-Chat-7B} & 44.78 & \textbf{{64.21}} & \textbf{{66.43}} & \textbf{{21.24}} &19.39 & 20.64 & 4.67 & 30.08 & 0 & 30.16 \\
    \quad X-KDE & \textbf{{47.67}} & 59.46 & 61.64 & 20.96 & \textbf{{29.36}} &\textbf{{32.63}} & \textbf{{45.13}} & \textbf{{30.75}} & \textbf{14.07} & \textbf{{37.96}} \\
    \bottomrule
    \end{tabular}
    }
    \caption{
        \textbf{General tasks performance of X-KDE, and LLaMA2-Chat-7B}. Results in \textbf{bold} represent the best performance in each category.
    }
    \label{tab:general_task}
\end{table*}

\paragraph{X-KDE can process thousands of edits simultaneously.} 

In line with the single-edit procedure, we evaluate both English and Chinese edits separately. For simplicity in presentation, we take the average of these two results, as shown in Figure~\ref{fig:batch_edit}.
As the batch size increases, we observe a gradual decline across all performance metrics for all method. The drop is particularly severe for MEMIT and FT-L, especially in the locality metric, which is nearly cut in half. In contrast, X-KDE achieves the best performance, maintaining the highest accuracy while exhibiting the slowest degradation rate. These results indicate that our approach remains stable in cross-lingual settings, even after thousands of edits.

\paragraph{X-KDE can sequential acquire new knowledge without forgetting previous information.} 

In the sequential-editing setting, the model integrates new knowledge on top of its previous edits, which leads to a gradual decline in performance over time.
As illustrated in Figure~\ref{fig:seq_edit}, the performance of methods that modify model parameters typically degrades as the number of edits increases. For instance, MEMIT and FT-L remain stable only when the number of edits $n\le 10$; beyond that, their performance deteriorates sharply. In contrast, knowledge storage-retrieval paradigms represented by X-KDE and LTE circumvent direct parameter modifications through external memory architectures. Moreover, X-KDE demonstrates superior cross-lingual transfer capabilities compared to LTE, achieving better performance across diverse data streams.

\subsection{Results of General Tasks} 

A series of studies have demonstrated that knowledge editing can influence model performance across various scenarios~\cite{yang2024butterfly,li2023unveiling,gu2024model}. To investigate whether our method impact the model's capabilities in unrelated domains, we conducted tests across a range of fields.
Given that the cross-lingual knowledge editing task typically involves two languages, we use English and Chinese as representative examples. Multiple benchmarks are selected in these two languages, covering tasks such as commonsense reasoning, natural language understanding, open-domain QA, and general intelligence.
For example, the benchmarks selected for English include MMLU~\cite{hendrycks2020measuring}, CommonSenseQA~\cite{talmor2018commonsenseqa}, PIQA~\cite{bisk2020piqa}, XSum~\cite{narayan2018don}, and Natural Questions~\cite{kwiatkowski2019natural}. For Chinese, the chosen benchmarks are CMMLU~\cite{li2023cmmlu}, CommonSenseQA\_zh~\cite{2023opencompass}, CEval~\cite{huang2024c}, and Natural Questions\_zh~\cite{2023opencompass}.
We conducted all experiments using the OpenCompass tool~\cite{2023opencompass}.
The results are presented in Table \ref{tab:general_task}. 
Overall, our method not only preserves the model's performance in English but also significantly enhances its capabilities in Chinese.
Although certain tasks, such as CommonsenseQA, XSum, and PIQA, show a decrease in performance when tested in English, the overall results demonstrate consistent English capabilities. This highlights the robustness of our method, which achieves cross-lingual knowledge editing while preserving the model's original performance and significantly improving its proficiency in Chinese.

\section{Analysis}
\subsection{Are both stages of X-KDE indispensable?}
We examine the significance of the two stages in the X-KDE method through our ablation experiments, as shown in Table~\ref{tab:stage}.
\begin{CJK*}{UTF8}{gbsn}
Focusing solely on the improvement in performance metrics, Stage 1 undoubtedly plays a decisive role in our method, achieving significant gains (up to +25.64\% average score) compared to the untrained baseline model. This stage enables cross-lingual knowledge editing via in-context learning, providing a substantial boost in model performance. While Stage 2 appears to offer a smaller improvement (a 2.25\% average gain), a closer analysis highlights its practical importance.
Stage 2 is particularly effective in adjusting the model’s output style to align with the target language, addressing issues such as incorrect language mixing (e.g., code-switching) or failure to generate responses in the expected linguistic format. For example, after Stage 1 updates the knowledge, the model may still produce a year like “2006” in the source language format. Stage 2 ensures the correct linguistic form, such as “2006年” in Chinese.
\end{CJK*}
The optimization of target-language preferences using ORPO not only improves factual accuracy but also ensures stylistic appropriateness in the target language. By refining the model’s preferences, ORPO helps it better adapt to the cultural and grammatical norms of the target language, addressing challenges like code-switching and maintaining consistency across multilingual contexts.

\begin{table}[ht]
\centering
\resizebox{1.0\linewidth}{!}{
\begin{tabular}{ccccc}
\toprule
\multirow{2}{*}{\textbf{Methods}} 
  & \multicolumn{2}{c}{\textbf{Stages}} 
  & \multicolumn{2}{c}{\textbf{Score}} \\
\cmidrule(lr){2-3} \cmidrule(lr){4-5}
 & \textbf{Stage-1} & \textbf{Stage-2} & \textbf{en-avg} & \textbf{zh-avg} \\
\midrule
Origin & \xmark & \xmark & 62.41 & 64.96 \\
\midrule
\multirow{3}{*}{X-KDE}
  & \cmark & \xmark & 88.05 & 86.99 \\
  & \xmark & \cmark & 77.14 & 66.38 \\
  & \cmark & \cmark & 91.04 & 88.49 \\
\bottomrule
\end{tabular}}
\caption{
\textbf{Ablation results on Bi-ZsRE benchmark} with Llama2-7b-chat as the base model. The \textit{en-avg} and \textit{zh-avg} columns denote average scores when editing in English or Chinese, respectively.
}
\label{tab:stage}
\end{table}

\subsection{Does every composition of the training data matter?}

In this section, we focus on the composition of the training data. As shown in Table~\ref{tab:ablation}, the absence of any specific segment of the training data leads to a noticeable decline in editing performance, whether in monolingual or cross-lingual settings.
Excluding either monolingual or cross-lingual training data causes a sharp drop in performance in the corresponding areas. When the in-scope data is omitted, the model tends to retain its original knowledge, resulting in reduced reliability, generality, and portability. On the other hand, removing the out-of-scope data causes the model to overly depend on the updated knowledge, thus diminishing locality. Similarly, removing the edit descriptors from the training data prevents the model from effectively utilizing the updated knowledge, leading to a drop in all metrics. Interestingly, training data with longer text samples seems to enhance the model’s comprehension and improve portability.
In summary, each component of the training data plays a unique and indispensable role, and omitting any part negatively impacts the model’s performance across all key metrics.

\begin{table}[H]
\centering
\resizebox{1.0\linewidth}{!}{
\begin{tabular}{lcccccccc}
\toprule
\multirow{2}{*}{\textbf{Methods}} & \multicolumn{4}{c}{\textbf{en-en}} & \multicolumn{4}{c}{\textbf{en-zh}} \\
\cmidrule(lr){2-5} \cmidrule(lr){6-9}
 & R. & G. & L. & P.  & R. & G. & L. & P. \\
\midrule
\textbf{X-KDE} & {99.93} & {99.87} & 90.15 & {76.41} & {94.81} & {94.65} & 95.05 & {77.43} \\
\midrule
-w/o mono. data 
& \textcolor{red}{78.93} &\textcolor{red}{76.60} & \textcolor{red}{77.33} & \textcolor{red}{68.21}
& 94.6 &94.52 &94.73 &76.02  \\
-w/o coss-lin. data 
& 99.91 & 99.81 & 88.97 & 77.40 & \textcolor{red}{76.86} & \textcolor{red}{76.82} & \textcolor{red}{86.99} & \textcolor{red}{67.49} \\
-w/o in-scope & \textcolor{red}{81.02} & \textcolor{red}{82.58} & 93.93 & \textcolor{red}{69.15} & \textcolor{red}{75.17} & \textcolor{red}{74.85} & 93.56 & \textcolor{red}{69.62} \\
-w/o out-of-scope & 99.99 & 99.45 & \textcolor{red}{70.71} & 76.64 & 92.91 & 92.73 & \textcolor{red}{76.21} & 73.48 \\
-w/o edit descriptor  & \textcolor{red}{87.53} & \textcolor{red}{81.99} & \textcolor{red}{67.69} & \textcolor{red}{66.53} & \textcolor{red}{84.26} & \textcolor{red}{84.12} & \textcolor{red}{79.43} & \textcolor{red}{74.13}  \\
-w/o long-text & 100 & 99.82 & 93.54 & \textcolor{red}{73.63} & 92.35 & 92.75 & 93.16 & \textcolor{red}{72.46} \\
\bottomrule
\end{tabular}}
\caption{
\textbf{Ablation results in the monolingual and the cross-lingual setting}, examining ``reliability'' (R.), ``generality'' (G.), ``locality'' (L.), and ``portability'' (P.).
}
\label{tab:ablation}
\end{table}

\subsection{Why choose ORPO as the preferred optimization method?}

We evaluate several popular preference optimization methods using a held-out dataset from our training data, specifically direct policy optimization (DPO)\cite{rafailov2023direct}, contrastive preference optimization (CPO)\cite{xu2024contrastive}, Kahneman-Tversky Optimization (KTO)\cite{ethayarajh2024kto}, and odds ratio preference optimization (ORPO)\cite{hong2403orpo} on the Bi-ZsRE benchmark. As shown in Table~\ref{tab:rl}, ORPO outperforms the other methods, achieving the best overall performance. While CPO and KTO also yield similar improvements, ORPO excels in preserving irrelevant target-language samples, demonstrating superior locality. In contrast, DPO results in a performance decline, which we attribute to the absence of negative log-likelihood (NLL) constraints, potentially weakening the model’s instruction-following capabilities. Based on these results, we adopt ORPO as the default optimization method for the second phase of our approach, as it provides the most significant improvement.

\begin{table}[H]
\centering
\resizebox{1.0\linewidth}{!}{
\begin{tabular}{lccccc}
\toprule
    \multirow{1}{*}{\textbf{Method}}
 & \textbf{Eff. }& \textbf{Gen.} &\textbf{ Loc.a}  & \textbf{Por.} & \underline{\textbf{Avg.}}  \\
\midrule

\textbf{SFT}  & \cellcolor{g1} 90.22 & \cellcolor{g1} 90.2 & \cellcolor{g1} 89.22 & \cellcolor{g1} 64.03 & \cellcolor{g1} \underline{83.41} \\
\textbf{~~~+DPO} & \cellcolor{r3} {88.47}  & \cellcolor{r2}{88.3}  & \cellcolor{r1} 89.18 & \cellcolor{r4} {61.11}  & \cellcolor{r3}\underline{83.26} \\
\textbf{~~~+CPO}  & \cellcolor{b3} 92.41 & \cellcolor{b3} 92.67 & \cellcolor{b1} 90.97 & \cellcolor{b3} 67.09 & \cellcolor{b3}\underline{85.78} \\
\textbf{~~~+KTO} & \cellcolor{b2} 92.23 & \cellcolor{b2} 92.01 & 89.22 & \cellcolor{b4} 67.23 & \cellcolor{b2}\underline{85.17} \\
\textbf{~~~+ORPO}  & \cellcolor{b4} \textbf{92.85} & \cellcolor{b4} \textbf{93.06} & \cellcolor{b4} \textbf{92.49} & \cellcolor{b4} \textbf{67.23} & \cellcolor{b4}\textbf{\underline{86.41}} \\
\bottomrule
\end{tabular}}
\caption{
\textbf{Effects of different preference optimization methods} with single edit setting on en-zh. Shades of cell color represent differences between preference optimization methods and simply SFT, where \textcolor{b4}{blue} denotes better performance while \textcolor{r4}{red} indicates worse.
}
\label{tab:rl}
\end{table}

\section{Related Work}

\paragraph{Knowledge Editing}  
The task of knowledge editing was introduced by ~\cite{sinitsin2020editable} to update specific knowledge while preserving unrelated information. Current methods fall into two paradigms: preserving or modifying the model’s parameters.
(1) \emph{Preserving LLMs' parameters} involves auxiliary models or extra parameters. SERAC~\cite{mitchell2022memory} uses a counterfactual model to update knowledge without altering model parameters. TPatcher~\cite{huang2023transformer} and CaliNET~\cite{dong2022calibrating} add trainable parameters to edit knowledge. IKE~\cite{zheng2023can} and ICE~\cite{cohen2024evaluating} leverage in-context learning to correct knowledge.
(2) \emph{Modifying the model's parameters} directly updates specific parameters to change knowledge. KE~\cite{de2021editing} and MEND~\cite{mitchell2021fast} predict weight updates for new data using a hyper-network. KN~\cite{dai2021knowledge}, ROME~\cite{meng2022rome}, and MEMIT~\cite{meng2022mass} use knowledge attribution or causal mediation analysis to target specific parameters for updating.

\paragraph{Cross-Lingual Knowledge Editing}  
Cross-lingual knowledge editing extends monolingual editing by propagating edits across languages. ~\cite{wang2023cross} introduced cross-lingual knowledge editing and created the Bi-ZsRE dataset to assess the applicability of monolingual methods in multilingual contexts. LiME~\cite{xu2022language} proposes language anisotropic editing to enhance cross-lingual editing, and MPN~\cite{si2024mpn} introduces multilingual patch neurons to update knowledge. However, these methods treat source language answers as ground truth for target language queries, falling short of achieving true cross-lingual transfer.

\paragraph{LLM Alignment}  
LLM alignment~\cite{gabriel2020artificial} ensures that LLMs' behaviors align with human values. Techniques such as supervised fine-tuning (SFT)~\cite{wei2021finetuned, wang2023far, mishra2021cross} train models to follow task descriptions in natural language. Despite SFT, models may still generate harmful content~\cite{carlini2021extracting, gehman2020realtoxicityprompts}. To address this, reinforcement learning with human feedback (RLHF)~\cite{stiennon2020learning, ouyang2022training} refines models further. Recent methods like SimPO~\cite{meng2024simpo} and ORPO~\cite{hong2403orpo} contribute to improving alignment in practical deployments.

\section{Conclusion}
\label{sec:conclusion}

In this paper, we present the Cross-Lingual Knowledge Democracy Edit (X-KDE) framework, which facilitates knowledge editing across languages in large language models (LLMs). By integrating Cross-lingual Edition Instruction Tuning (XE-IT) and Target-language Preference Optimization (TL-PO), X-KDE efficiently transfers knowledge from a source language to a target language while maintaining strong performance in monolingual settings. Additionally, we introduce high-quality datasets specifically designed for cross-lingual knowledge editing, filling gaps in existing resources. Our experimental results demonstrate that X-KDE outperforms current methods, offering a scalable solution for cross-lingual knowledge editing. Future research will explore applying X-KDE to other domains and optimizing its efficiency.

\section*{Limitations}
While our work presents promising results, there are a few limitations to consider. First, due to computational constraints, we validate X-KDE on models with up to 7B parameters. Evaluating larger models, such as those exceeding 70B parameters, could provide more robust insights. Second, while our method has been effective in multilingual settings, its application to additional domains, such as finance or law, remains unexplored. Future research will focus on scaling X-KDE and extending its applicability to other fields.
\section*{Ethics and Reproducibility Statements}
\paragraph{Ethics}  
We take ethical considerations seriously and strictly adhere to the ACL Ethics Policy. All datasets used in this work are publicly available and widely adopted by the research community. Our methods focus on enhancing the multilingual capabilities of large language models without introducing harmful biases or unethical content. We ensure that all experiments are conducted in compliance with ethical guidelines, prioritizing fairness and transparency in model deployment.

\paragraph{Reproducibility}  
In this paper, we discuss the detailed experimental setup, including training hyperparameters, baseline implementations, and statistical descriptions. More importantly, \textit{\textbf{we have provided our code and data in the Supplementary Material}} to help reproduce the experimental results of this paper. Due to space limitations during uploading, the full dataset will be released upon acceptance.

\section*{Acknowledgements}
We are grateful to the anonymous reviewers and the area chair for their insightful comments and suggestions. Dr Tao’s research is partially supported by NTU RSR and Start Up Grants.

\normalem
\bibliography{custom}
\appendix
\onecolumn

\section{Details of Dataset Construction}
\subsection{Prompt Details of Sample Generation}
\label{sec:appendix_sample}
Here, we present the detailed prompts for sample generation. Specifically, we guide LLMs in producing the queries and answers via the following prompts:

\begin{tcolorbox}
[colback=lightgray!20,colframe=darkgray!80,title= Query Generation Prompt]
\label{tab:quality_prompt}
[Edit description] will modify the knowledge. You must think according to the narrative of [Edit description].
\newline
\newline

[Edit description]: Who is Chris Klemmer affiliated with? University of Washington

[Prompt]: Please generate a question related to Chris Klemmer. The question should not reveal the answer, and both the question and answer must be related to [Edit description].

[Generated Question]: Which university is Chris Klemmer associated with?
\newline
\newline
[Edit description]: What profession does Dagmar Lurz pursue? Film director

[Prompt]: Please generate a question related to Dagmar Lurz. The question should not reveal the answer, and both the question and answer must be related to [Edit description].

[Generated Question]: In which creative field has Dagmar Lurz displayed her extraordinary talent?
\newline
\newline
[Edit description]: When was Jana Fesslová born? April 20, 1977

[Prompt]: Please generate a question related to Jana Fesslová. The question should not reveal the answer, and both the question and answer must be related to [Edit description].

[Generated Question]: What major historical or cultural event in the 1970s aligns with Jana Fesslová's birth date?
\newline
\newline
[Edit description]: What type of voice does Martin Crosby have? Contralto

[Prompt]: Please generate a question related to Martin Crosby. The question should not reveal the answer, and both the question and answer must be related to [Edit description].

[Generated Question]: What is Martin Crosby's vocal type, and how does this type bring unique timbre and expressiveness to musical performances?
\newline
\newline
[Edit description]: Who directed "The Mentor"? Jean-Luc Godard

[Prompt]: Please generate a question related to "The Mentor". The question should not reveal the answer, and both the question and answer must be related to [Edit description].

[Generated Question]: Who directed "The Mentor," and how did they use their unique directorial style to shape the film's artistic quality and influence?
\newline
\newline
\newline
[Edit description]: \texttt{<instruct>}
\newline
[Prompt]: Please generate a question related to \texttt{<subject>}. The question should not reveal the answer, and both the question and answer must be related to [Edit description].
\newline
[Generated Question]: \texttt{<answer>}

\end{tcolorbox}

\begin{tcolorbox}
[colback=lightgray!20,colframe=darkgray!80,title= Answer Generation Prompt]
\label{tab:quality_prompt}
Imagine that the description of [Edit description] is correct. Please answer [question] based on [Edit description].
\newline
\newline

[Edit description]: Who is Chris Klemmer affiliated with? University of Washington

[Question]: Who is Chris Klemmer affiliated with?

[Prompt]: The answer should focus on Chris Klemmer, with University of Washington as a secondary focus, avoiding unnecessary details.

[Answer]: Chris Klemmer is affiliated with the University of Washington. During his academic career at the University of Washington, he made significant achievements and contributions:

1.Academic Research: Klemmer conducted extensive research at the University of Washington, particularly in his field of expertise. His research findings not only advanced academic knowledge but also had a considerable impact on practical applications.

2.Teaching Contributions: As a professor at the University of Washington, Klemmer excelled in teaching and mentored many outstanding students. His teaching methods and course designs were well-received and inspired students to innovate.

3.Academic Leadership: Klemmer held various academic leadership positions at the University of Washington, such as department chair or research center director. In these roles, he promoted academic projects, fostered interdisciplinary collaboration, and enhanced the institution's academic reputation.

4.Publications: Klemmer published numerous high-level academic papers and books, which received widespread attention and citation in the academic community, further solidifying his standing in academia.
\newline
\newline
[Edit description]: What profession does Dagmar Lurz pursue? Film director

[Question]: In which field has Dagmar Lurz shown exceptional talent?

[Prompt]: The answer should focus on Dagmar Lurz, with film directing as secondary, avoiding unrelated content.

[Answer]: Dagmar Lurz has demonstrated her exceptional talent in the field of film directing. As a film director, she is responsible for guiding actors' performances and overseeing the film's overall artistic style and narrative pacing. Her work involves interpreting scripts, arranging scenes, guiding cinematography, and making editing decisions, all requiring rich creativity and sharp artistic insight. Through her directorial work, Dagmar Lurz has left a unique mark on film art, offering viewers deep visual and emotional experiences.
\newline
\newline
\newline
\newline
[Edit description]: \texttt{<instruct>}
\newline
[Question]: \texttt{<question>}
\newline
[Prompt]: Please generate a question related to \texttt{<subject>}. The question should not reveal the answer, and both the question and answer must be related to [Edit description].
The answer should focus on \texttt{<subject>}, with \texttt{<new answer>} as secondary, avoiding unrelated content.
\newline
[Generated Answer]: \texttt{<new answer>}
\end{tcolorbox}

\begin{tcolorbox}
[colback=lightgray!20,colframe=darkgray!80,title= Out-of-scope Generation Prompt]
\label{tab:quality_prompt}
In the following statements, "changed answer" represents the modified factual knowledge. When the answer is changed, other properties of the subject should remain unchanged. For example, if we edit basketball player Grant Hill to a football player, this would not affect his nationality. Therefore, for irrelevant attributes such as country, the output should remain consistent with the pre-edit version. You should recall an irrelevant attribute and generate a question and answer based on that irrelevant attribute and the "subject".
\newline
\newline
Question: Who is the father of Juan María Bordaberry?

Subject: Juan María Bordaberry

Changed answer: Gabrielle Bordaberry

Irrelevant attribute recalled: place of death

New question: Where did Juan María Bordaberry die?

New answer: Montevideo
\newline
\newline
Question: Who published the game Street Rod 2?

Subject: Street Rod 2

Changed answer: Sierra Entertainment

Irrelevant attribute recalled: release format

New question: What is the release format of Street Rod 2?

New answer: Floppy disk
\newline
\newline
Question: What is the status of the Cross River Gorilla?

Subject: Cross River Gorilla

Changed answer: Endangered

Irrelevant attribute recalled: classification level

New question: What is the classification level of the Cross River Gorilla?

New answer: Subspecies"""
\newline
\newline
\newline
[Question]: \texttt{<question>}
\newline
[Subject]: \texttt{<subject>}
\newline
[Changed Answer]: \texttt{<new answer>}
\end{tcolorbox}

\subsection{Prompt Details of Quality Control}
\label{sec:appendix_quality}
Moreover, we employ LLMs in Quality Controls. Judging and scoring via the following prompts:

\begin{tcolorbox}
[colback=lightgray!20,colframe=darkgray!80, title= Prompt for judging ]
\label{tab:quality_prompt}
Please act as a fair judge and determine whether the [answer] answers the [question] based on the [Edit description]. Provide an explanation and strictly follow the format:
\newline
- If the answer is based on the edit description, output "[T]"
\newline
- If it is not, output "[F]".
\newline
\newline
[Edit description]: \texttt{<instruct>}
\newline
[Question]: \texttt{<question>}
\newline
[Answer]: \texttt{<answer>}
\end{tcolorbox}

\begin{tcolorbox}
[colback=lightgray!20,colframe=darkgray!80,title= Prompt for scoring ]
\label{tab:quality_prompt}
Please act as a fair judge and rate the sentence based on the following criteria:
\newline
1. Sentence complexity: Evaluate the complexity of the sentence, such as inversion, imperative sentences, sentences with word inflections, or sentences starting with multiple adverbs, nouns, and subjects. The more complex, the higher the score.
\newline
2. Vocabulary richness: Evaluate the diversity of vocabulary used. The more diverse, the higher the score.
\newline
3. Faithfulness: Evaluate whether the [answer] faithfully adheres to [Edit description], meaning it accurately answers [question]. If the answer highly matches the description, the score is higher; if the question leaks the answer, deduct points.
\newline
Provide separate scores for each criterion (1-10), and calculate the average score, then output the sentence with the highest score. The output format should be:
\newline
[Sentence complexity: score; Vocabulary richness: score; Faithfulness: score; Overall score: score]
\newline
\newline
\newline
[Edit description]: \texttt{<instruct>}
\newline
[Question]: \texttt{<question>}
\newline
[Answer]: \texttt{<answer>}
\end{tcolorbox}

\subsection{Training Data Statistics}
\label{sec:appendix_data_construction}
Table \ref{tab:statistics} lists the statistics of our constructed high-quality dataset, which includes 388k samples and covers two languages: English (En) and Chinese (Zh). To prevent data leakage, we only sample instances from the training sets.

\section{Experimental Setup}
\subsection{Alignment Score Calculation}
\label{sec:aligenment}
We employ NLLB-600M-distilled\footnote{\url{https://huggingface.co/facebook/nllb-200-distilled-600M}} as the multilingual translation model. And since the translation model is trained on a large-scale parallel dataset in both source and target languages ($X$ and $Y$) by maximizing the conditional generation probability $P( Y | X ; \theta)$. Higher alignment between $Y$ and $X$ results in a higher conditional generation probability $P(Y | X)$. To compute the "alignment" score, we input the model's response and force-decode the ground-truth $Y$ in target language.

\subsection{Implementation Details}
\label{sec:appendix_implementation}
All experiments were executed on 8 NVIDIA A100 GPUs (80G). We employ EasyEdit~\cite{wang2023easyedit} to implement all the baselines with the default settings. We employ llamafactory~\cite{zheng2024llamafactory} to implement Cross-lingual Edition Instruction Tuning (XE-IT) phase of our method. When training on the English editing only subset, the duration is approximately 10 to 15 hours. Hyper-parameters of our X-KDE are in Table \ref{tab:hyperparameters}.

\begin{table}[!h]
\small
\centering
{\begin{tabular}{lcc}
\toprule
\textbf{Hyperparameter} & \textbf{XE-IT} & \textbf{TL-PO} \\ 
\midrule
Learning rate           & 5e-6             & 1e-6              \\
Max length              & 2560             & 1024              \\
Optimizer               & AdamW            & AdamW             \\
Scheduler               & cosine           & cosine            \\
Weight decay            & 0.1              & 0.05                \\
warmup steps            & 100              & 100   
\\ \bottomrule
\end{tabular}}
\caption{\label{tab:hyperparameters}
\textbf{Hyper-parameters} for training our X-KDE.}
\end{table}

\section{Used Scientific Artifacts}
\label{sec:Scientific}
We list scientific artifacts used in our work blow.  And we use of these existing artifacts is consistent with their intended use.
\begin{itemize} [itemsep=1pt]
    
    \item \textit{DeepSpeed (Apache-2.0 license)}~\footnote{ \url{https://github.com/deepspeedai/DeepSpeed}}, a deep learning optimization library to improve the efficiency of training large language models.
    \item \textit{Transformers (Apache-2.0 license)}~\footnote{ \url{https://github.com/huggingface/transformers}}, a framework that provides state-of-the-art pretrained models for NLP tasks
    \item \textit{trl (Apache-2.0 license)}~\footnote{ \url{https://github.com/huggingface/trl}}, a library designed to integrate reinforcement learning (RL) with transformer models.
    \item \textit{vLLM (Apache-2.0 license)}~\footnote{ \url{https://github.com/vllm-project/vllm}}, an optimized framework for inference with large language models.
    
\end{itemize}

\begin{table*}[!t]
\footnotesize
\centering
\resizebox{0.8\linewidth}{!}{
\begin{tabular}{lccc ccc ccc}
\toprule
\multirow{2}{*}{\textbf{Data Source}} & \multirow{2}{*}{\textbf{Lang.}} & \multicolumn{2}{c}{\textbf{\# in-scope}} & \multicolumn{2}{c}{\textbf{\# out-scope}} & \multirow{2}{*}{\textbf{\# Total}} & \multirow{2}{*}{\textbf{Avg Token}} \\ 
\cmidrule(lr){3-4} \cmidrule(lr){5-6}
 & & w/ edit & w/o edit & w/ edit & w/o edit \\

\midrule
\multirow{2}{*}{ZsRE}       & En  & 20,000                     & 20,000                      & 20,000                         & 20,000                          & 80,000       &  48        \\
& Zh  & 20,000                     & 20,000                      & 20,000                         & 20,000                          & 80,000       &  84       \\

\multirow{2}{*}{HalluEditBench} & En  & 2,000 & 2,000 & 2,000     & 2,000  & 8,000       &  38         \\
& Zh  & 2,000 & 2,000 & 2,000 & 2,000  & 8,000  &  60         \\

\multirow{2}{*}{RIPPLEEDITS} & En & 2,250                     & 2,250                      & 2,250                         & 2,250                          & 9,000       & 53          \\
& Zh & 2,250                     & 2,250                      & 2,250                         & 2,250                          & 9,000       & 88          \\

\multirow{2}{*}{WikiBio}  & En   & 250                       & 250                        & 250                           & 250                            & 1,000       & 162          \\
& Zh   & 250                       & 250                        & 250                           & 250                            & 1,000       & 294          \\

\multirow{2}{*}{MQUAKE}    & En  & 4,000                     & 4,000                      & 4,000                         & 4,000                          & 16,000      &  266       \\
& Zh  & 4,000                     & 4,000                      & 4,000                         & 4,000                          & 16,000      &  334         \\

\multirow{2}{*}{COUNTERFACT} & En & 7,500                     & 7,500                      & 7,500                         & 7,500                          & 30,000      & 530          \\ 
& Zh & 7,500                     & 7,500                      & 7,500                         & 7,500                          & 30,000      & 888          \\ \midrule
\multirow{2}{*}{Total}   & En    & 36,000                    & 36,000                    & 36,000                 & 36,000                       & 144,000    & 170 \\
& Zh    & 36,000                    & 36,000                    & 36,000                 & 36,000                       & 144,000    & 266 \\
\bottomrule
\end{tabular}
}
\caption{\textbf{Statistics of our training data (Lang.:language)}. ``Avg Token'' denotes the average length(token-level) of samples, and ``edit'' indicates the edit descriptor.}
\label{tab:statistics}
\end{table*}

\section{Supplemental Experiment Results} 
\label{sec:appendix_experiment}

\subsection{Detailed Results of Llama2-7b-chat}
\label{sec:appendix_ex_llama2}
More detailed results on MzsRE of Llama2-7b-chat are listed in Table~\ref{tab:ap-en-edit} and Table~\ref{tab:ap-zh-edit}.

\subsection{Detailed Results of Qwen2.5-7B-Instruct}
\label{sec:appendix_ex_qwen2}
To verify the universality of our method, we conducted experiments on Qwen2.5-7B-Instruct. More detailed results are listed in Table~\ref{table:qwen-res}, Table~\ref{tab:qw-en-edit} and Table~\ref{tab:qw-zh-edit}.

\begin{table*}[!h]
    \centering
    \scalebox{0.7}{
        \begin{tabular}{ccccccccccccccc}
        \toprule
         \textbf{Metrics} & \textbf{Methods}  & \textbf{en-en} & \textbf{en-cz} & \textbf{en-de} & \textbf{en-du} & \textbf{en-es} & \textbf{en-fr} & \textbf{en-pt} & \textbf{en-ru} & \textbf{en-th} & \textbf{en-tr} & \textbf{en-vi} & \textbf{en-zh} & \underline{\textbf{en-avg}}\\
         \midrule
           \multirow{7}{*}{{\textbf{Reliability}}} 
           & FT-L &  52.92 &  41.81 &  39.79 &  39.02 &  39.49 &  39.72 &  39.26 &  39.79 &  36.44 &  36.86 &  46.21 &  51.81 &  \underline{41.93} \\
           & FT-M &  99.96 &  66.93 &  70.16 &  67.17 &  63.69 &  64.98 &  64.22 &  48.96 &  36.46 &  57.54 &  66.80 &  56.89 &  \underline{63.65} \\
           & ROME  &  96.36 &  56.54 &  60.82 &  58.89 &  57.41 &  56.43 &  54.91 &  41.69 &  35.44 &  45.76 &  56.94 &  49.94 &  \underline{55.93} \\
           & MEMIT &  95.44 &  62.37 &  64.82 &  64.12 &  59.46 &  61.90 &  58.69 &  44.54 &  36.40 &  49.15 &  61.34 &  52.05 &  \underline{59.19} \\
           & IKE & 99.65 & 83.22 & 80.61 & 79.36 & 76.69 & 78.48 & 75.37 & 67.62 & 54.38 & 76.90 & 81.22 & 67.83 & \underline{76.78} \\
           & LTE  & \textcolor{darkgreen}{100.00} & 84.29 & 81.71 & 80.60 & 77.67 & 79.11 & 77.39 & 72.02 & 62.04 & 78.87 & 81.92 & 76.93 & \underline{79.38} \\
           \cmidrule{2-15}
           & X-KDE & 99.93 & \textcolor{darkgreen}{92.78} & \textcolor{darkgreen}{87.43} & \textcolor{darkgreen}{88.89} & \textcolor{darkgreen}{85.71} & \textcolor{darkgreen}{87.49} & \textcolor{darkgreen}{89.87} & \textcolor{darkgreen}{89.32} & \textcolor{darkgreen}{89.66} & \textcolor{darkgreen}{91.23} & \textcolor{darkgreen}{87.55} & \textcolor{darkgreen}{93.07} & \underline{\textcolor{darkgreen}{90.24}} \\
           \midrule
            \multirow{7}{*}{\textbf{Generality}} 
            & FT-L &  49.60 &  40.75 &  38.87 &  38.36 &  39.68 &  39.12 &  39.56 &  38.97 &  36.89 &  37.18 &  45.89 &  51.71 &  \underline{41.38} \\
            & FT-M &  95.53 &  65.45 &  68.15 &  65.09 &  62.39 &  62.28 &  61.63 &  47.69 &  36.88 &  56.87 &  65.97 &  56.52 &  \underline{62.04} \\
            & ROME &  85.13 &  54.99 &  58.91 &  56.99 &  56.58 &  54.47 &  53.94 &  40.68 &  35.36 &  45.06 &  56.38 &  50.31 &  \underline{54.07} \\
            & MEMIT &  89.59 &  60.71 &  63.80 &  61.98 &  58.10 &  59.40 &  57.63 &  43.31 &  36.77 &  48.68 &  60.51 &  52.01 &  \underline{57.71} \\
            & IKE & 99.54 & 82.67 & 80.78 & 79.18 & 76.37 & 78.22 & 75.49 & 67.51 & 54.26 & 76.97 & 80.99 & 67.88 & \underline{76.65} \\
            & LTE & \textcolor{darkgreen}{99.87} & 84.26 & 81.63 & 81.07 & 77.51 & 78.99 & 77.38 & 71.46 & 61.90 & 78.26 & 81.37 & 76.24 & \underline{79.16} \\
           \cmidrule{2-15}
           & X-KDE & 99.68 & \textcolor{darkgreen}{92.87} & \textcolor{darkgreen}{87.25} & \textcolor{darkgreen}{88.87} & \textcolor{darkgreen}{85.16} & \textcolor{darkgreen}{87.57} & \textcolor{darkgreen}{89.93} & \textcolor{darkgreen}{89.10} & \textcolor{darkgreen}{89.21} & \textcolor{darkgreen}{91.25} & \textcolor{darkgreen}{87.62} & \textcolor{darkgreen}{93.11} & \underline{\textcolor{darkgreen}{90.14}} \\
           \midrule
           \multirow{7}{*}{\textbf{Locality}} 
           
            & FT-L &  93.96 &  90.78 &  81.06 &  88.98 &  83.32 &  89.30 &  90.98 &  89.53 &  90.18 &  88.95 &  93.02 &  85.56 &  \underline{88.80} \\
            & FT-M &  97.71 &  96.94 &  96.24 &  96.57 &  96.36 &  97.56 &  97.49 &  97.31 &  96.93 &  97.25 &  98.04 &  94.61 &  \underline{96.92} \\
            & ROME &  97.81 &  96.12 &  97.57 &  96.80 &  97.36 &  96.98 &  97.14 &  96.70 &  96.28 &  96.83 &  97.60 &  97.70 &  \underline{97.07} \\
            & MEMIT &  \textcolor{darkgreen}{98.55} &  \textcolor{darkgreen}{98.24} &  \textcolor{darkgreen}{98.55} &  \textcolor{darkgreen}{98.08} &  \textcolor{darkgreen}{98.35} &  \textcolor{darkgreen}{98.30} &  \textcolor{darkgreen}{98.45} &  \textcolor{darkgreen}{98.33} &  \textcolor{darkgreen}{98.88} &  \textcolor{darkgreen}{98.97} &  \textcolor{darkgreen}{98.89} &  \textcolor{darkgreen}{98.76} &  \underline{\textcolor{darkgreen}{98.53}} \\
            & IKE & 58.13 & 61.35 & 65.57 & 61.52 & 63.93 & 60.42 & 59.42 & 58.90 & 68.84 & 63.97 & 68.40 & 64.54 & \underline{62.91 }\\
            & LTE & 89.28 & 77.01 & 77.90 & 77.68 & 81.54 & 81.51 & 81.23 & 78.39 & 79.86 & 76.34 & 82.93 & 86.63 & \underline{80.86} \\
           \cmidrule{2-15}
           & X-KDE & 93.12 & 78.76 & 79.88 & 77.19 & 81.29 & 78.97 & 80.00 & 82.78 & 82.08 & 72.62 & 82.11 & 91.91 & \underline{81.73} \\ 
           \midrule
           \multirow{7}{*}{\textbf{Portability}} 
           & FT-L &  52.85 &  46.85 &  43.51 &  43.21 &  44.47 &  44.91 &  43.72 &  47.05 &  39.92 &  41.14 &  54.05 &  55.13 &  \underline{46.40} \\
           & FT-M &  57.17 &  48.66 &  46.38 &  46.34 &  47.09 &  47.54 &  46.36 &  48.41 &  38.55 &  42.50 &  55.53 &  52.16 &  \underline{48.06} \\    
           & ROME &  58.46 &  49.79 &  48.58 &  47.06 &  48.29 &  48.83 &  47.30 &  49.21 &  38.11 &  42.38 &  56.62 &  51.81 &  \underline{48.87} \\
           & MEMIT &  57.02 &  50.41 &  47.96 &  47.26 &  47.26 &  48.47 &  47.21 &  49.25 &  38.56 &  43.44 &  57.16 &  52.19 &  \underline{48.85} \\
           & IKE & 70.97 & 56.44 & 58.87 & 56.91 & 58.05 & 58.41 & 56.33 & 57.96 & 39.46 & 48.69 & 62.86 & 58.97 & \underline{56.99} \\
           & LTE & \textcolor{darkgreen}{77.29} & \textcolor{darkgreen}{61.85} & \textcolor{darkgreen}{64.91} & \textcolor{darkgreen}{63.82} & \textcolor{darkgreen}{61.53} & \textcolor{darkgreen}{62.83} & \textcolor{darkgreen}{62.39} & \textcolor{darkgreen}{61.43} & 44.51 & \textcolor{darkgreen}{51.04} & \textcolor{darkgreen}{65.37} & 67.47 & \underline{\textcolor{darkgreen}{62.04}} \\

           \cmidrule{2-15}
           & X-KDE & 76.13 & 60.53 & 58.74 & 56.94 & 55.19 & 58.85 & 58.81 & 62.89 & \textcolor{darkgreen}{56.18} & 48.69 & 61.78 & \textcolor{darkgreen}{74.04} & \underline{60.73} \\
           \bottomrule
        \end{tabular}
    }
    \caption{
    \textbf{Results on MzsRE dataset for editing performed in English} using Llama2-7b-chat. Here, ``en-zh'' means that English serves as the source language and Chinese as the target language, with similar interpretations for the other pairs. \underline{``en-avg''} denotes the average performance across cross-lingual scenarios.}
    \label{tab:ap-en-edit}
\end{table*}

\begin{table*}[!h]
    \centering
    \scalebox{0.7}{
        \begin{tabular}{ccccccccccccccc}
        \toprule
         \textbf{Metrics} & \textbf{Methods}  & \textbf{zh-en} & \textbf{zh-cz} & \textbf{zh-de} & \textbf{zh-du} & \textbf{zh-es} & \textbf{zh-fr} & \textbf{zh-pt} & \textbf{zh-ru} & \textbf{zh-th} & \textbf{zh-tr} & \textbf{zh-vi} & \textbf{zh-zh} & \underline{\textbf{zh-avg}}\\
         \midrule
           \multirow{7}{*}{\textbf{Reliability}} 
            & FT-L & 40.81 & 38.16 & 36.21 & 35.60 & 36.28 & 36.45 & 35.55 & 38.88 & 33.98 & 34.50 & 43.32 & 54.79 & \underline{38.71} \\
            & FT-M & 51.87 & 48.51 & 46.71 & 45.70 & 45.69 & 45.98 & 45.65 & 46.97 & 39.44 & 44.74 & 54.06 & \textcolor{darkgreen}{100.00} & \underline{51.28} \\
            & ROME & 44.15 & 40.06 & 38.04 & 37.62 & 38.44 & 37.99 & 37.69 & 39.25 & 32.94 & 36.49 & 44.82 & 73.48 & \underline{41.75} \\
            & MEMIT & 51.87 & 41.45 & 39.61 & 39.29 & 39.19 & 39.23 & 38.78 & 40.85 & 33.77 & 38.49 & 46.72 & 76.12 & \underline{43.78} \\
            & IKE & 65.88 & 68.68 & 67.63 & 66.75 & 65.06 & 65.63 & 63.82 & 63.52 & 52.39 & 61.32 & 70.90 & 99.85 & \underline{67.62} \\
            & LTE & 65.44 & 64.74 & 62.05 & 62.91 & 61.09 & 60.85 & 61.20 & 63.09 & 55.71 & 58.15 & 67.02 & 99.76 & \underline{65.17} \\
           \cmidrule{2-15}
            & X-KDE & \textcolor{darkgreen}{94.64} & \textcolor{darkgreen}{84.40} & \textcolor{darkgreen}{83.05} & \textcolor{darkgreen}{81.08} & \textcolor{darkgreen}{80.33} & \textcolor{darkgreen}{81.22} & \textcolor{darkgreen}{83.38} & \textcolor{darkgreen}{82.56} & \textcolor{darkgreen}{83.09} & \textcolor{darkgreen}{78.69} & \textcolor{darkgreen}{81.47} & 99.99 & \underline{\textcolor{darkgreen}{84.49}} \\
           \midrule
           \multirow{7}{*}{\textbf{Generality}} 
            & FT-L & 40.67 & 37.70 & 36.35 & 35.18 & 36.60 & 35.49 & 35.67 & 38.19 & 33.86 & 34.97 & 43.14 & 53.90 & \underline{38.48} \\
            & FT-M & 51.24 & 48.24 & 46.49 & 45.30 & 45.71 & 45.63 & 45.81 & 46.32 & 39.62 & 45.06 & 54.42 & \textcolor{darkgreen}{99.68} & \underline{51.13} \\
            & ROME & 43.80 & 39.72 & 38.01 & 37.83 & 38.26 & 36.74 & 38.62 & 38.46 & 32.76 & 36.46 & 45.02 & 71.13 & \underline{41.40} \\
            & MEMIT & 51.24 & 41.16 & 40.14 & 38.22 & 39.10 & 38.80 & 39.06 & 40.15 & 34.18 & 38.22 & 46.34 & 74.21 & \underline{43.40} \\
            & IKE & 65.75 & 67.90 & 67.48 & 66.39 & 65.01 & 65.35 & 63.50 & 63.44 & 52.72 & 61.03 & 70.27 & 99.28 & \underline{67.34} \\
            & LTE & 64.94 & 64.53 & 62.72 & 62.31 & 61.15 & 60.50 & 61.11 & 62.94 & 55.39 & 58.29 & 66.88 & 98.69 & \underline{64.95} \\
           \cmidrule{2-15}
            & X-KDE & \textcolor{darkgreen}{94.51} & \textcolor{darkgreen}{84.27} & \textcolor{darkgreen}{82.43} & \textcolor{darkgreen}{81.46} & \textcolor{darkgreen}{80.12} & \textcolor{darkgreen}{81.08} & \textcolor{darkgreen}{82.69} & \textcolor{darkgreen}{82.19} & \textcolor{darkgreen}{82.81} & \textcolor{darkgreen}{78.33} & \textcolor{darkgreen}{81.07} & 98.89 & \underline{\textcolor{darkgreen}{84.15}} \\
           \midrule
           \multirow{7}{*}{\textbf{Locality}} 
            & FT-L & 94.81 & 89.42 & 83.41 & 88.81 & 83.09 & 89.92 & 90.09 & 86.63 & 79.94 & 85.75 & 89.69 & 66.38 & \underline{85.66} \\
            & FT-M & \textcolor{darkgreen}{98.19} & 96.70 & 92.82 & 95.79 & 93.90 & 97.68 & 97.17 & 96.46 & 93.05 & 95.49 & 97.18 & 79.74 & \underline{94.51} \\
            & ROME & 97.93 & 96.17 & 97.41 & 96.23 & 96.81 & 96.66 & 96.99 & 95.89 & 94.36 & 96.15 & 97.23 & 96.42 & \underline{96.52} \\
            & MEMIT & \textcolor{darkgreen}{98.19} & \textcolor{darkgreen}{98.05} & \textcolor{darkgreen}{98.52} & \textcolor{darkgreen}{98.68} & \textcolor{darkgreen}{98.76} & \textcolor{darkgreen}{98.52} & \textcolor{darkgreen}{98.50} & \textcolor{darkgreen}{97.51} & \textcolor{darkgreen}{96.55} & \textcolor{darkgreen}{98.10} & \textcolor{darkgreen}{98.36} & \textcolor{darkgreen}{96.06} & \underline{\textcolor{darkgreen}{97.98}} \\
            & IKE & 69.41 & 63.74 & 63.22 & 64.02 & 62.24 & 61.69 & 62.10 & 61.15 & 67.15 & 64.73 & 69.31 & 67.91 & \underline{64.72} \\
            & LTE & 89.26 & 76.59 & 80.09 & 78.01 & 81.86 & 81.44 & 81.22 & 80.31 & 80.24 & 76.63 & 82.92 & 86.67 & \underline{81.27 }\\
           \cmidrule{2-15}
            & X-KDE & 94.07 & 80.53 & 81.79 & 78.13 & 83.49 & 81.18 & 81.72 & 84.22 & 84.14 & 75.66 & 83.15 & 92.20 & \underline{83.36 }\\
           \midrule
           \multirow{7}{*}{\textbf{Portability}} 
            & FT-L & 55.25 & 47.54 & 45.76 & 44.73 & 46.21 & 46.61 & 44.96 & 48.20 & 37.99 & 41.16 & 54.13 & 48.07 & \underline{46.72} \\
            & FT-M & 55.30 & 48.13 & 45.78 & 45.25 & 46.09 & 47.04 & 45.22 & 48.75 & 40.32 & 42.12 & 55.57 & 61.79 & \underline{48.45} \\
            & ROME & 52.66 & 47.19 & 44.60 & 44.89 & 44.56 & 45.64 & 44.11 & 47.96 & 36.40 & 40.23 & 53.87 & 48.34 & \underline{45.87} \\
            & MEMIT & 55.30 & 47.78 & 45.99 & 45.60 & 46.00 & 46.91 & 45.32 & 48.10 & 37.82 & 41.51 & 54.88 & 50.83 & \underline{47.17} \\
            & IKE & 63.06 & 54.77 & 58.40 & 56.25 & 55.59 & 56.72 & 54.58 & 57.65 & 40.80 & 47.15 & 61.58 & 66.44 & \underline{56.08} \\
            & LTE & \textcolor{darkgreen}{68.30} & 59.86 & \textcolor{darkgreen}{60.92} & \textcolor{darkgreen}{59.62} & \textcolor{darkgreen}{58.27} & \textcolor{darkgreen}{59.71} & 58.59 & 60.03 & 45.83 & 48.53 & \textcolor{darkgreen}{63.11} & 69.40 & \underline{59.35} \\
           \cmidrule{2-15}
            & X-KDE & 67.95 & \textcolor{darkgreen}{60.34} & 59.10 & 58.10 & 55.89 & 58.43 & \textcolor{darkgreen}{59.37} & \textcolor{darkgreen}{61.73} & \textcolor{darkgreen}{56.27} & \textcolor{darkgreen}{49.69} & 62.78 & \textcolor{darkgreen}{73.43} & \underline{\textcolor{darkgreen}{60.26}} \\
           \bottomrule
        \end{tabular}
    }
    \caption{\textbf{Results on MzsRE dataset for editing performed in Chinese} using Llama2-7b-chat. Here, ``zh-en'' means that Chinese serves as the source language and English as the target, with similar interpretations for the other pairs. \underline{``zh-avg''} denotes the average performance across cross-lingual scenarios.}
    \label{tab:ap-zh-edit}
\end{table*}

\begin{table*}[t]
    \centering
    \resizebox*{0.9\linewidth}{!}{
    \begin{tabular}{cccccccccc}
    \toprule
    \multicolumn{1}{c}{\multirow{2}{*}{\bf Method}} & \multicolumn{4}{c}{\textbf{Test in English}} & \multicolumn{4}{c}{\textbf{Test in Chinese}} & \\
    \cmidrule(lr){2-5} \cmidrule(lr){6-9} 
  \multicolumn{1}{c}{} & \textbf{Reliability} & \textbf{Generality} & \textbf{Locality} & \textbf{Portability} & \textbf{Reliability} & \textbf{Generality} & \textbf{Locality} & \textbf{Portability} & \textbf{\textit{\underline{Avg.}}} \\
    \midrule
\multicolumn{10}{c}{\textbf{Edit in English}} \\
\toprule
FT-L & 62.89 & 65.93 & 75.06 & 38.83 & 44.25 & 44.36 & 68.35 & 40.21 & \underline{54.98} \\
FT-M & \textcolor{darkgreen}{100.0} & 99.35 & 92.44 & 52.59 & 64.03 & 63.61 & 88.48 & 53.22 & \underline{76.72} \\
ROME & 99.48 & 92.83 & 98.63 & 56.69 & 58.91 & 58.44 & 98.47 & 54.93 & \underline{77.30} \\ 
MEMIT & 96.77 & 89.30 & \textcolor{darkgreen}{98.78} & 55.60 & 59.97 & 59.35 & \textcolor{darkgreen}{98.49} & 54.31 & \underline{76.57} \\
IKE & 97.32 & 98.56 & 50.26 & 67.30 & 69.89 & 69.45 & 57.07 & 57.94 & \underline{70.97} \\
LTE & 99.78 & 99.28 & 87.64 & \textcolor{darkgreen}{74.23} & 73.95 & 74.40 & 84.34 & 61.85 & \underline{81.93} \\ 
\cmidrule{1-10}
X-KDE(Ours) & 99.72 & \textcolor{darkgreen}{99.56} & 88.79 & 73.96 & \textcolor{darkgreen}{90.42} & \textcolor{darkgreen}{90.20} & 91.53 & \textcolor{darkgreen}{62.59} & \underline{\textcolor{darkgreen}{87.10}} \\  
    \midrule
    
    \multicolumn{10}{c}{\textbf{Edit in Chinese}} \\
    \toprule
FT-L & 32.59 & 33.17 & 73.39 & 37.89 & 48.85 & 53.49 & 54.88 & 28.17 & \underline{45.30} \\
FT-M & 53.31 & 52.80 & 92.53 & 51.52 & \textcolor{darkgreen}{100.0} & \textcolor{darkgreen}{99.85} & 79.38 & 52.84 & \underline{72.78} \\
ROME & 45.66 & 45.31 & 98.31 & 52.74 & 99.36 & 94.77 & 97.97 & 57.06 & \underline{73.90} \\
MEMIT & 45.68 & 44.25 & \textcolor{darkgreen}{98.94} & 52.26 & 98.07 & 94.20 & \textcolor{darkgreen}{96.69} & 57.70 & \underline{73.47} \\
IKE & 79.59 & 78.77 & 49.57 & 65.20 & 96.37 & 96.47 & 66.05 & 61.59 & \underline{74.20} \\
LTE & 79.40 & 78.50 & 86.64 & \textcolor{darkgreen}{70.24} & 98.95 & 98.60 & 84.54 & \textcolor{darkgreen}{64.53 }& \underline{82.68}\\
\cmidrule{1-10}
X-KDE(Ours) & \textcolor{darkgreen}{94.78} & \textcolor{darkgreen}{94.77} & 95.14 & 67.50 & 99.79 & 98.29 & 90.57 & 61.30 & \underline{\textcolor{darkgreen}{87.77}} \\
    \midrule
    \end{tabular}
}
    \caption{
    \textbf{Cross-lingual editing performance of different methods} on Qwen2.5-7B-Instruct backbones. Results in \textcolor{darkgreen}{green} indicates the best results. ``\underline{\textbf{\textit{Avg.}}}'' represents the overall mean of all metrics evaluated across the two languages.
    }
    \label{table:qwen-res}
\end{table*}

\begin{table*}[!h]
    \centering
    \scalebox{0.7}{
        \begin{tabular}{ccccccccccccccc}
        \toprule
         \textbf{Metrics} & \textbf{Methods}  & \textbf{en-en} & \textbf{en-cz} & \textbf{en-de} & \textbf{en-du} & \textbf{en-es} & \textbf{en-fr} & \textbf{en-pt} & \textbf{en-ru} & \textbf{en-th} & \textbf{en-tr} & \textbf{en-vi} & \textbf{en-zh} & \underline{\textbf{en-avg}}\\
         \midrule
           \multirow{7}{*}{{\textbf{Reliability}}} 
           & FT-L & 63.77 & 50.88 & 50.30 & 47.23 & 47.40 & 51.08 & 48.18 & 42.59 & 44.37 & 47.86 & 48.39 & 44.91 & \underline{48.91} \\
           & FT-M & \textcolor{darkgreen}{100.0} & 71.56 & 74.67 & 70.47 & 66.45 & 68.94 & 68.82 & 57.35 & 51.57 & 67.36 & 64.79 & 64.59 & \underline{68.88} \\
           & ROME  & 99.44 & 55.82 & 63.27 & 61.38 & 57.41 & 59.44 & 60.23 & 49.78 & 48.74 & 53.73 & 53.06 & 59.23 & \underline{60.13} \\
           & MEMIT & 96.92 & 54.87 & 61.36 & 58.79 & 54.67 & 58.15 & 58.02 & 49.53 & 47.71 & 52.49 & 51.88 & 60.27 & \underline{58.72} \\
           & IKE & 97.89 & 82.71 & 82.84 & 78.05 & 76.34 & 79.26 & 78.69 & 69.95 & 66.43 & 77.28 & 75.93 & 70.53 & \underline{77.99} \\
           & LTE  & 99.70 & 84.28 & 84.73 & 80.76 & 78.02 & 82.07 & 80.40 & 76.52 & 69.53 & 81.06 & 77.89 & 73.35 & \underline{80.69} \\
           \cmidrule{2-15}
           & X-KDE & 98.56 & \textcolor{darkgreen}{89.60 }& \textcolor{darkgreen}{86.69} & \textcolor{darkgreen}{86.94} &\textcolor{darkgreen}{ 84.84 }&\textcolor{darkgreen}{ 84.87} &\textcolor{darkgreen}{ 85.30} & \textcolor{darkgreen}{89.05 }& \textcolor{darkgreen}{87.61 }& \textcolor{darkgreen}{90.00 }& \textcolor{darkgreen}{79.47 }& \textcolor{darkgreen}{89.74} & \underline{\textcolor{darkgreen}{87.72}} \\
           \midrule
            \multirow{7}{*}{\textbf{Generality}} 
            & FT-L & 66.81 & 50.62 & 50.42 & 46.74 & 47.63 & 51.18 & 48.35 & 42.85 & 44.81 & 47.66 & 47.50 & 44.91 & \underline{49.12} \\
            & FT-M &  99.26 & 70.46 & 73.67 & 69.37 & 65.68 & 67.06 & 66.14 & 56.27 & 51.83 & 65.35 & 62.55 & 64.09 & \underline{67.64} \\
            & ROME & 93.67 & 54.71 & 61.05 & 58.82 & 55.48 & 57.48 & 58.01 & 48.59 & 48.32 & 51.49 & 51.57 & 59.13 & \underline{58.19} \\
            & MEMIT & 90.32 & 54.69 & 59.12 & 56.52 & 53.93 & 55.96 & 55.53 & 48.25 & 47.77 & 51.16 & 50.64 & 59.72 & \underline{56.97} \\
            & IKE & 98.52 & 82.75 & 82.71 & 77.83 & 75.86 & 78.92 & 78.30 & 69.57 & 66.83 & 77.26 & 75.26 & 70.68 & \underline{77.87} \\
            & LTE & \textcolor{darkgreen}{99.30} & 84.48 & 84.57 & 80.39 & 77.93 & 81.49 & 80.34 & 76.56 & 69.45 & 81.07 & 77.56 & 73.63 & \underline{80.56 }\\
           \cmidrule{2-15}
           & X-KDE & 98.00 &\textcolor{darkgreen}{ 89.91} &\textcolor{darkgreen}{ 86.55} &\textcolor{darkgreen}{ 87.07} & \textcolor{darkgreen}{84.88} & \textcolor{darkgreen}{84.53} & \textcolor{darkgreen}{85.49} & \textcolor{darkgreen}{89.03} & \textcolor{darkgreen}{87.49} & \textcolor{darkgreen}{89.82} & \textcolor{darkgreen}{79.42} & \textcolor{darkgreen}{89.95} & \underline{\textcolor{darkgreen}{87.68}} \\
           \midrule
           \multirow{7}{*}{\textbf{Locality}} 
           
            & FT-L & 74.68 & 75.09 & 62.23 & 69.32 & 62.46 & 70.98 & 73.56 & 74.63 & 79.92 & 66.58 & 75.33 & 67.88 & \underline{71.06} \\
            & FT-M &  92.46 & 90.64 & 83.27 & 87.98 & 83.72 & 90.23 & 90.39 & 90.89 & 93.08 & 86.20 & 91.12 & 88.30 & \underline{89.02} \\
            & ROME & 98.71 & 97.43 & 97.32 & 98.05 & 98.35 & 97.37 & 98.41 & 98.49 & 98.46 & 97.53 & \textcolor{darkgreen}{98.53} & 98.35 & \underline{98.08} \\
            & MEMIT & \textcolor{darkgreen}{98.76} &\textcolor{darkgreen}{ 98.49} & \textcolor{darkgreen}{98.21 }&\textcolor{darkgreen}{ 98.49 }& \textcolor{darkgreen}{98.77 }& \textcolor{darkgreen}{98.63 }& \textcolor{darkgreen}{98.48} &\textcolor{darkgreen}{ 98.74 }&\textcolor{darkgreen}{ 98.75} & \textcolor{darkgreen}{98.59} & 98.47 & \textcolor{darkgreen}{98.51} &\underline{\textcolor{darkgreen}{ 98.57}} \\
            & IKE & 50.52 & 55.71 & 57.39 & 53.66 & 57.51 & 58.40 & 56.95 & 61.17 & 65.64 & 59.46 & 60.03 & 57.45 & \underline{57.82} \\
            & LTE & 88.28 & 80.06 & 81.15 & 78.18 & 82.89 & 84.62 & 82.80 & 86.32 & 85.12 & 78.61 & 80.00 & 84.74 & \underline{82.73} \\
           \cmidrule{2-15}
           & X-KDE & 95.22 & 77.82 & 87.16 & 77.71 & 79.86 & 83.98 & 82.35 & 87.42 & 86.40 & 75.54 & 80.53 & 92.59 & \underline{83.88} \\
           \midrule
           \multirow{7}{*}{\textbf{Portability}} 
           & FT-L & 38.08 & 38.77 & 36.92 & 36.49 & 37.42 & 39.98 & 40.22 & 45.27 & 47.89 & 39.07 & 35.92 & 40.23 & \underline{39.69 }\\
           & FT-M &  52.17 & 49.42 & 48.36 & 46.27 & 46.92 & 50.18 & 49.86 & 53.84 & 54.22 & 46.93 & 45.66 & 52.83 & \underline{49.72} \\   
           & ROME & 56.40 & 49.68 & 49.33 & 48.00 & 49.17 & 53.04 & 51.94 & 54.93 & 54.53 & 47.56 & 46.14 & 54.95 & \underline{51.31} \\
           & MEMIT & 54.82 & 49.69 & 49.62 & 47.80 & 49.48 & 51.74 & 51.67 & 55.01 & 54.57 & 48.15 & 46.94 & 54.32 & \underline{51.15} \\
           & IKE & 67.00 & 55.75 & 59.25 & 55.11 & 56.31 & 60.00 & 58.76 & 59.04 & 55.32 & 52.44 & 52.16 & 58.27 & \underline{57.45} \\
           & LTE  &\textcolor{darkgreen}{ 74.25 }& \textcolor{darkgreen}{61.72 }& \textcolor{darkgreen}{65.09 }& \textcolor{darkgreen}{61.27} & \textcolor{darkgreen}{62.63 }&\textcolor{darkgreen}{ 65.92} & \textcolor{darkgreen}{64.21} & \textcolor{darkgreen}{65.30 }& \textcolor{darkgreen}{60.78} &\textcolor{darkgreen}{ 59.96 }& \textcolor{darkgreen}{57.94} & \textcolor{darkgreen}{61.36} &\underline{\textcolor{darkgreen}{ 63.37}} \\

           \cmidrule{2-15}
           & X-KDE & 70.36 & 53.71 & 54.93 & 52.23 & 54.76 & 56.20 & 56.57 & 60.27 & 58.51 & 52.41 & 49.23 & 61.18 & \underline{56.70} \\
           \bottomrule
        \end{tabular}
    }
    \caption{
    \textbf{Results on MzsRE dataset for editing performed in English} using Qwen2.5-7B-Instruct. Here, ``en-zh'' means that English serves as the source language and Chinese as the target language, with similar interpretations for the other pairs. ``en-avg'' denotes the average performance across cross-lingual scenarios.}
    \label{tab:qw-en-edit}
\end{table*}

\begin{table*}[!h]
    \centering
    \scalebox{0.7}{
        \begin{tabular}{ccccccccccccccc}
        \toprule
         \textbf{Metrics} & \textbf{Methods}  & \textbf{zh-en} & \textbf{zh-cz} & \textbf{zh-de} & \textbf{zh-du} & \textbf{zh-es} & \textbf{zh-fr} & \textbf{zh-pt} & \textbf{zh-ru} & \textbf{zh-th} & \textbf{zh-tr} & \textbf{zh-vi} & \textbf{zh-zh} & \underline{\textbf{zh-avg}}\\
         \midrule
           \multirow{7}{*}{\textbf{Reliability}} 
            & FT-L & 33.24 & 34.37 & 32.47 & 31.74 & 30.58 & 33.23 & 32.99 & 35.73 & 40.39 & 29.41 & 28.90 & 48.62 & \underline{34.31} \\
            & FT-M & 52.97 & 51.26 & 51.88 & 49.22 & 48.13 & 49.82 & 51.21 & 52.25 & 51.33 & 50.00 & 47.02 & 100.00 & \underline{54.59} \\
            & ROME & 46.29 & 42.41 & 43.29 & 42.32 & 41.15 & 42.64 & 43.13 & 45.94 & 47.14 & 44.22 & 40.61 & 99.51 & \underline{48.22} \\
            & MEMIT& 46.36 & 42.81 & 44.02 & 42.29 & 40.87 & 42.69 & 43.75 & 46.06 & 46.78 & 43.18 & 41.53 & 98.59 & \underline{48.25 }\\
            & IKE & 80.22 & 71.25 & 76.32 & 69.69 & 69.28 & 71.10 & 70.64 & 66.51 & 65.03 & 68.96 & 66.66 & 96.32 & \underline{72.67} \\
            & LTE  & 79.54 & 71.83 & 75.40 & 71.18 & 67.28 & 70.02 & 68.88 & 72.39 & 66.11 & 70.15 & 68.79 & 99.03 & \underline{73.38} \\
           \cmidrule{2-15}
            & X-KDE& \textcolor{darkgreen}{94.97} &\textcolor{darkgreen}{ 79.86 }& \textcolor{darkgreen}{84.24} & \textcolor{darkgreen}{79.12 }& \textcolor{darkgreen}{77.97 }& \textcolor{darkgreen}{81.24} &\textcolor{darkgreen}{ 79.16} &\textcolor{darkgreen}{ 75.77 }& \textcolor{darkgreen}{67.75 }& \textcolor{darkgreen}{77.76} & \textcolor{darkgreen}{76.48} &\textcolor{darkgreen}{ 99.85 }& \underline{\textcolor{darkgreen}{81.18 }}\\
           \midrule
           \multirow{7}{*}{\textbf{Generality}} 
            & FT-L & 32.90 & 34.32 & 32.80 & 31.56 & 31.00 & 33.88 & 33.40 & 35.69 & 40.79 & 29.89 & 29.57 & 53.08 & \underline{34.91} \\
            & FT-M & 52.66 & 51.13 & 52.15 & 49.12 & 47.66 & 48.95 & 50.46 & 51.47 & 51.51 & 50.22 & 46.62 &  99.96 & \underline{54.33} \\
            & ROME & 46.11 & 42.31 & 42.60 & 41.57 & 40.65 & 41.88 & 42.77 & 44.60 & 46.34 & 42.97 & 40.80 & 95.23 & \underline{47.32} \\
            & MEMIT & 45.03 & 42.58 & 42.79 & 41.50 & 40.83 & 41.81 & 42.51 & 45.20 & 46.48 & 42.65 & 41.33 & 95.06 & \underline{47.31} \\
            & IKE & 78.68 & 72.05 & 76.01 & 69.73 & 69.72 & 71.07 & 71.09 & 66.08 & 65.18 & 69.40 & 66.86 & 96.50 & \underline{72.70} \\
            & LTE & 78.20 & 71.15 & 75.18 & 70.72 & 66.41 & 69.68 & 69.22 & 72.32 & 66.04 & 69.93 & 68.50 & 98.34 & \underline{72.97} \\
           \cmidrule{2-15}
            & X-KDE  & \textcolor{darkgreen}{94.93} & \textcolor{darkgreen}{79.44 }& \textcolor{darkgreen}{84.67 }& \textcolor{darkgreen}{79.08 }& \textcolor{darkgreen}{77.85 }& \textcolor{darkgreen}{81.37} &\textcolor{darkgreen}{ 79.20} & \textcolor{darkgreen}{75.49} &\textcolor{darkgreen}{ 68.11 }& \textcolor{darkgreen}{77.53 }& \textcolor{darkgreen}{76.29} & \textcolor{darkgreen}{98.51 }& \underline{\textcolor{darkgreen}{81.04}} \\
           \midrule
           \multirow{7}{*}{\textbf{Locality}} 
            & FT-L & 73.25 & 67.27 & 58.50 & 63.17 & 58.05 & 65.53 & 68.75 & 67.52 & 69.27 & 56.25 & 64.19 & 54.75 & \underline{63.87} \\
            & FT-M & 92.53 & 89.49 & 83.74 & 87.61 & 85.73 & 89.50 & 90.45 & 88.96 & 90.36 & 83.69 & 90.04 &  79.06 & \underline{87.60} \\
            & ROME & 98.26 & 97.35 & 97.07 & 96.83 & 97.80 & 97.37 & 97.90 & 97.94 & 97.92 & 96.74 & 97.54 &\textcolor{darkgreen}{ 97.90} & \underline{97.55} \\
            & MEMIT &\textcolor{darkgreen}{ 98.94} & \textcolor{darkgreen}{98.36} & \textcolor{darkgreen}{98.42 }&\textcolor{darkgreen}{ 97.94} & \textcolor{darkgreen}{98.49 }& \textcolor{darkgreen}{98.69 }& \textcolor{darkgreen}{98.81} &\textcolor{darkgreen}{ 98.36} & \textcolor{darkgreen}{97.69 }& \textcolor{darkgreen}{97.96} & \textcolor{darkgreen}{97.76} & 96.67 & \underline{\textcolor{darkgreen}{98.18}} \\
            & IKE & 50.43 & 57.50 & 57.28 & 53.59 & 55.05 & 56.53 & 56.76 & 61.92 & 68.25 & 61.89 & 61.34 & 65.49 & \underline{58.84 }\\
            & LTE & 87.33 & 79.75 & 81.18 & 76.95 & 83.05 & 83.28 & 80.42 & 85.36 & 84.31 & 78.27 & 80.59 & 84.56 & \underline{82.09} \\
           \cmidrule{2-15}
            & X-KDE & 93.87 & 81.90 & 84.70 & 81.11 & 86.56 & 87.89 & 87.14 & 88.08 & 90.18 & 79.84 & 84.63 & 90.42 & \underline{86.36} \\
           \midrule
           \multirow{7}{*}{\textbf{Portability}} 
            & FT-L & 37.03 & 36.69 & 34.82 & 34.70 & 34.97 & 37.81 & 37.40 & 41.47 & 44.90 & 33.05 & 30.24 & 27.71 & \underline{35.90} \\
            & FT-M & 50.78 & 47.32 & 45.81 & 44.66 & 45.38 & 47.77 & 47.23 & 51.59 & 53.59 & 44.78 & 42.89 &  51.91 & \underline{47.81} \\
            & ROME & 51.70 & 47.48 & 47.36 & 45.42 & 46.54 & 50.09 & 49.16 & 54.37 & 54.39 & 46.86 & 44.10 & 56.78 & \underline{49.52} \\
            & MEMIT & 51.54 & 47.74 & 47.22 & 45.78 & 46.19 & 49.21 & 48.87 & 54.48 & 54.24 & 46.25 & 44.32 & 57.65 & \underline{49.46} \\
            & IKE & 64.32 & 56.14 & 58.98 & 55.44 & 56.10 & 59.75 & 58.63 & 61.12 & 57.28 & 53.05 & 52.57 & 61.75 & \underline{57.93} \\
            & LTE & \textcolor{darkgreen}{70.11} & \textcolor{darkgreen}{61.02} & \textcolor{darkgreen}{64.45} & \textcolor{darkgreen}{60.44} &\textcolor{darkgreen}{ 62.24} & \textcolor{darkgreen}{65.28 }& \textcolor{darkgreen}{63.82} & \textcolor{darkgreen}{65.28} &\textcolor{darkgreen}{ 60.67} & \textcolor{darkgreen}{59.51} & \textcolor{darkgreen}{57.18 }& \textcolor{darkgreen}{64.36} & \underline{\textcolor{darkgreen}{62.86}} \\
           \cmidrule{2-15}
            & X-KDE & 67.44 & 56.98 & 61.07 & 58.20 & 60.59 & 61.93 & 61.62 & 63.60 & 58.95 & 55.87 & 55.28 & 60.92 & \underline{60.20}\\
           \bottomrule
        \end{tabular}
    }
    \caption{\textbf{Results on MzsRE dataset for editing performed in Chinese} using Qwen2.5-7B-Instruct. Here, ``zh-en'' means that Chinese serves as the source language and English as the target, with similar interpretations for the other pairs. ``zh-avg'' denotes the average performance across cross-lingual scenarios.}
    \label{tab:qw-zh-edit}
\end{table*}




\end{document}